\documentclass{article}

\PassOptionsToPackage{numbers, square}{natbib}
\PassOptionsToPackage{hyphens}{url}

\usepackage[main,final]{neurips_2026}

\usepackage[utf8]{inputenc} 
\usepackage[T1]{fontenc}    
\usepackage{hyperref}       
\usepackage{url}            
\usepackage{booktabs}       
\usepackage{amsfonts}       
\usepackage{nicefrac}       
\usepackage{microtype}      
\usepackage{graphicx}
\usepackage{xcolor}         
\usepackage{scalerel}
\usepackage{fontawesome5}
\usepackage{amsmath}
\usepackage[noabbrev,capitalize,nameinlink]{cleveref}
\crefname{appendix}{Appendix}{Appendices}
\Crefname{appendix}{Appendix}{Appendices}
\usepackage{caption}
\usepackage{wrapfig}
\usepackage{listings}
\usepackage[most,skins,theorems]{tcolorbox}
\usepackage{algpseudocode}
\usepackage{mathtools}
\usepackage{multirow}
\usepackage{colortbl}
\algnewcommand\algorithmicconstants{\textbf{Constants:}}
\algnewcommand\Constants{\item[\algorithmicconstants]}

\DeclareCaptionType{alg}[Algorithm][List of Algorithms]
\crefname{alg}{Algorithm}{Algorithms}
\Crefname{alg}{Algorithm}{Algorithms}
\crefname{equation}{Equation}{Equations}
\Crefname{equation}{Equation}{Equations}
\crefformat{equation}{Eq.~#2#1#3}

\definecolor{tablecolor}{rgb}{0.8,0.8,0.8}

\newcommand{\ci}[1]{\textsubscript{\,#1}}
\newcommand{\valc}[2]{{#1}\ci{#2}}

\definecolor{editblue}{rgb}{0.0,0.0,0.75}

\newcommand{\qwenlogo}{\raisebox{-0.1\height}{\includegraphics[height=2em]{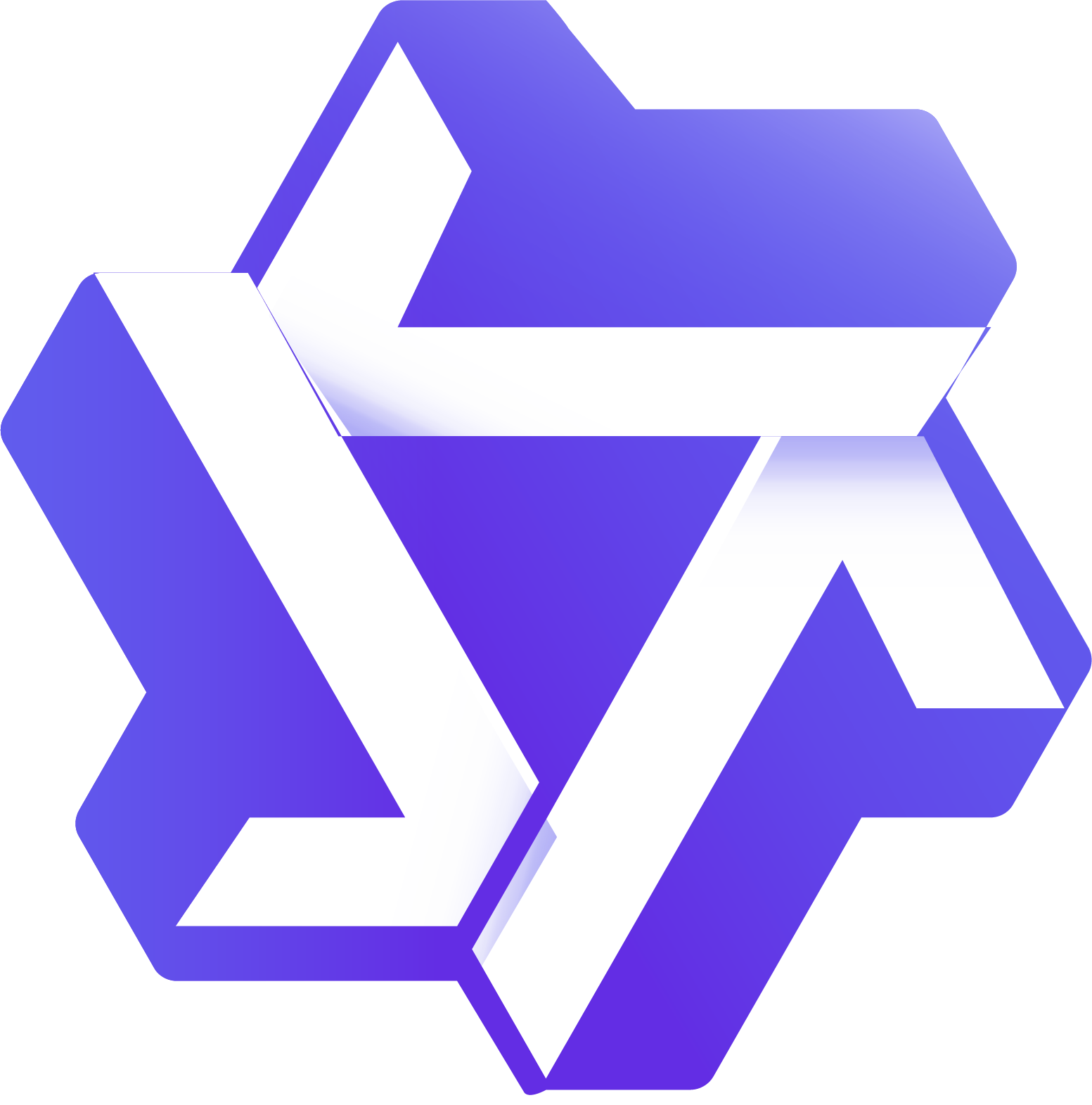}}}
\newcommand{\olmologo}{\raisebox{-0.1\height}{\includegraphics[height=2em]{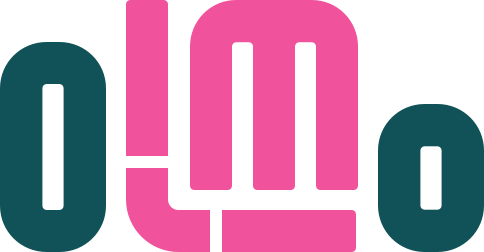}}}

\newcommand{\cg}{\cellcolor{gray!30}}
\newcommand{\ukp}{\raisebox{-0.25\height}{\includegraphics[height=1em]{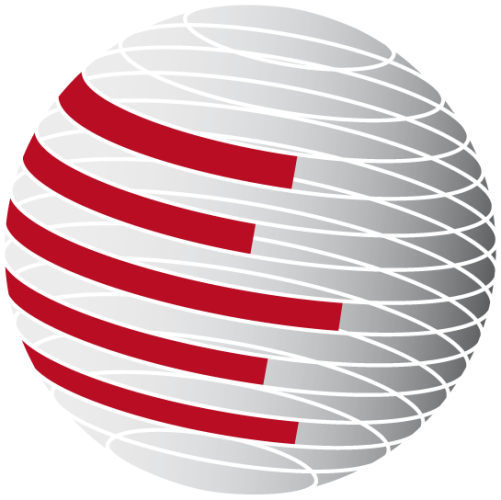}}}
\newcommand{\insait}{\raisebox{-0.25\height}{\includegraphics[height=1em]{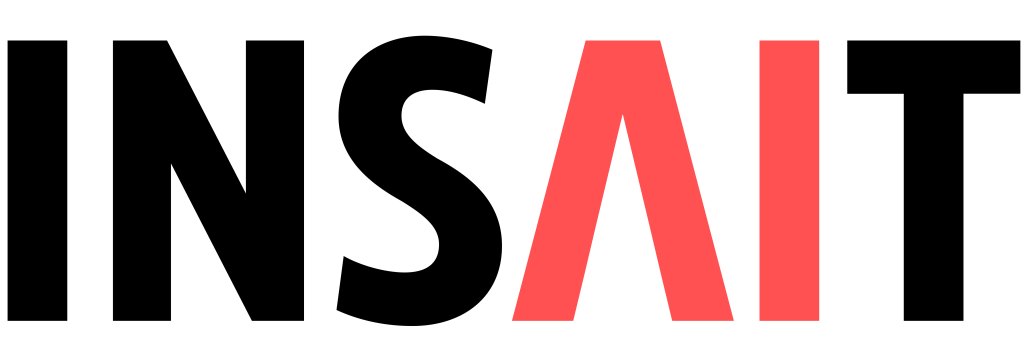}}}
\newcommand{\method}{3PO}
\newcommand{\methodfullname}{{Perturbed Parameter Policy Optimization}}
\newcommand{\batch}{B3PO}
\newcommand{\interleaved}{C3PO}
\newcommand{\mcsamples}{M3PO}

\newcommand{\vparam}{\boldsymbol{\theta}}

\newcommand{\dkl}[3]{\mathbb{D}_{\text{KL}}^{#1}(#2 \, \|\, #3)}

\newcommand\cut[1]{}

\newcommand{\squishlist}{
    \begin{list}{$\bullet$}
    { \setlength{\itemsep}{0pt}      \setlength{\parsep}{3pt}
        \setlength{\topsep}{3pt}       \setlength{\partopsep}{0pt}
        \setlength{\leftmargin}{1.5em} \setlength{\labelwidth}{1em}
        \setlength{\labelsep}{0.5em} } }

\newcommand{\squishlisttwo}{
    \begin{list}{$\bullet$}
    { \setlength{\itemsep}{0pt}    \setlength{\parsep}{0pt}
        \setlength{\topsep}{0pt}     \setlength{\partopsep}{0pt}
        \setlength{\leftmargin}{2em} \setlength{\labelwidth}{1.5em}
        \setlength{\labelsep}{0.5em} } }

\newcommand{\squishend}{
    \end{list}  }

{}
{}
{}

\newcommand{\myexpect}{\mathbb{E}}

\newcommand{\myvec}[1]{\mbox{$\mathbf{#1}$}}
\newcommand{\myvecsym}[1]{\mbox{$\boldsymbol{#1}$}}

\newcommand{\vsigma}{\mbox{$\myvecsym{\sigma}$}}

\newcommand{\vh}{\mbox{$\myvec{h}$}}

\newcommand{\vm}{\mathbf{m}}

\newcommand{\vr}{\mbox{$\myvec{r}$}}

\newcommand{\vx}{\mbox{$\myvec{x}$}}

\newcommand{\vy}{\mbox{$\myvec{y}$}}

\newcommand{\calQ}{\mbox{${\cal Q}$}}

\newcommand{\data}{\mathcal{D}}

\newcommand{\vocab}{\mathcal{V}}

\title{Parameter Exploration for RLVR via \\ Variational Learning}

\author{
Vatsal Venkatkrishna\,\textsuperscript{\insait{}\,\faEnvelope},
~Nico Carlsen Daheim\,\textsuperscript{\ukp{}},
~Iryna Gurevych\,\textsuperscript{\insait{}\,\ukp{}}
\\
\textsuperscript{\insait{}}\,INSAIT, Sofia University ``St. Kliment Ohridski'', Bulgaria\\
\textsuperscript{\ukp{}}\,Ubiquitous Knowledge Processing Lab (UKP Lab), \\
                         Department of Computer Science, Technical University of Darmstadt and \\
                        National Research Center for Applied Cybersecurity ATHENE, Germany \\
\\
{\raisebox{-0.25\height}{\includegraphics[height=1.2em]{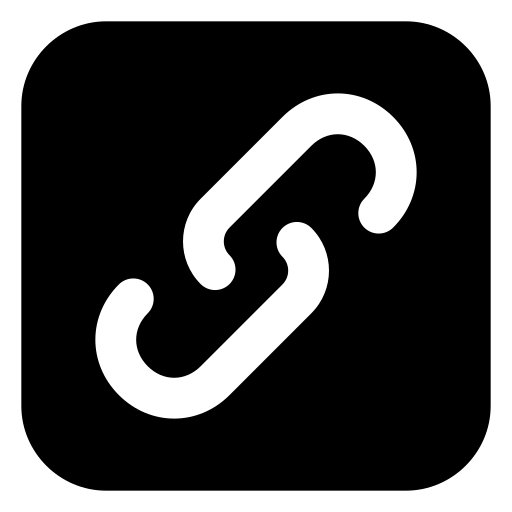}}} \texttt{\href{https://c3po.insait.ai/}{c3po.insait.ai}} \\
}

\begin{document}

\maketitle{
    \def\thefootnote{}\footnotetext{\faEnvelope \hspace{.2em} Corresponding author: \texttt{vatsal.venkatkrishna@insait.ai}}
}

\begin{abstract}
    Exploration has been a focus of reinforcement learning research for a long time. Recently, there has been growing evidence that it is also an important ingredient in LLM reinforcement learning recipes that can significantly impact downstream performance. Many existing methods control exploration in the action-space, for example, using temperature scaling. However, these methods cannot reorder tokens but only influence the variance in the output distribution. This limits exploration and can lead to divergence or stalled training. Here, we investigate parameter-space exploration, where rollouts are generated by sampling different policies from a posterior that may each explore different rollouts. Sampling less or more diverse policies is then a complementary control lever over exploration.
    We introduce a family of methods called Perturbed Parameter Policy Optimization (3PO) which use different sampling strategies and different rollout grouping for reward estimation.
    Experiments on Olmo-3-1025-7B and Qwen2.5-Math-7B across mathematical reasoning and code generation tasks show that these approaches can improve the average downstream performance over standard GRPO at a near-identical FLOPs cost. Moreover, using multiple parameter samples generally produces fewer zero-advantage groups and malformed or incorrect rollouts during training than GRPO and action-space baselines. Overall, our work presents evidence that parameter-space exploration can improve reinforcement learning for LLMs.


\end{abstract}

\section{Introduction}
Reinforcement Learning with Verifiable Rewards (RLVR) is a powerful post-training paradigm that can be used to improve the reasoning capabilities of LLMs without relying on human demonstrations~\citep{DBLP:journals/corr/abs-2411-15124,DBLP:journals/corr/abs-2402-03300}. Broadly, reinforcement learning could discover novel solutions by learning from autonomous experience~\citep{DBLP:books/lib/SuttonB98,DBLP:journals/corr/abs-2109-00157}, but current RLVR algorithms are frequently limited by a model's capabilities prior to RL~\citep{chen2026exploration,DBLP:journals/corr/abs-2504-07912,DBLP:conf/aaai/WuZDXZJFZLZFLZZ26}. Since the model learns exclusively from self-generated rollouts, high-reward trajectories may simply have too low probability and are thus often not discovered~\citep{DBLP:journals/corr/abs-2507-14843}.
This is particularly problematic in algorithms with grouped advantage calculations like Group-Relative Policy Optimization (GRPO)~\citep{DBLP:journals/corr/abs-2402-03300}, where there is no learning signal whenever all rollouts for a prompt receive the same reward. Such zero-advantage groups stall learning and therefore increase the already high compute costs associated with RLVR~\citep{noukhovitch2025faster,DBLP:journals/corr/abs-2601-12186}.
\begin{figure}[!t]
    \centering
    \includegraphics[height=4.5cm]{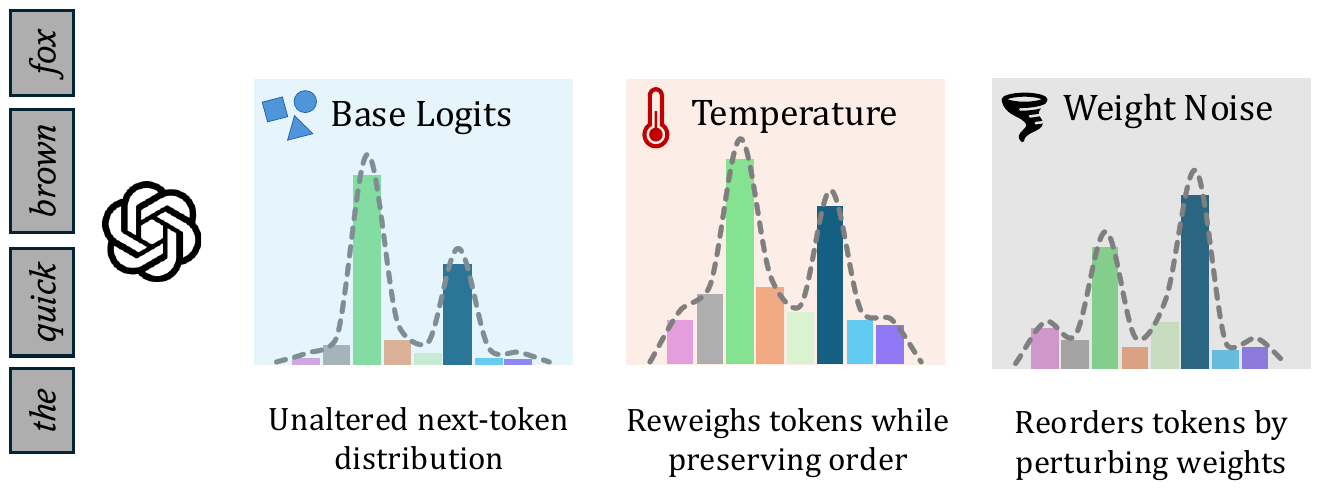}
    \caption{\textbf{Temperature versus Weight noise.} Temperature scaling controls the entropy of the token distribution but preserves relative token ordering. Adding noise to the model's weights can completely alter the token distribution, helping the model explore new regions of the solution space.}
    \label{fig: teaser}
\end{figure}

A common solution is to increase the softmax temperature during rollout generation to promote exploration~\citep{Polaris2025,DBLP:journals/corr/abs-2510-08892}. However, as illustrated in \cref{fig: teaser}, temperature can only control the variance of the token distribution and not the relative token order, which limits the trajectories that can be explored. Moreover, it uniformly alters the probability of all tokens in the vocabulary, which can lead to degeneration at high temperatures when erroneous tokens become likely~\citep{DBLP:journals/corr/abs-1904-09751}. Other complementary strategies like advantage shaping~\citep{DBLP:journals/corr/abs-2506-01939,zhu2025the,yu2025dapo} and entropy regularization~\citep{DBLP:conf/iclr/0003LKD21,DBLP:journals/corr/abs-2509-25133,DBLP:journals/corr/abs-2509-03493} have been proposed, but rather aim to stabilize training than promote exploration.

Parameter-space exploration has a long track record in more classical RL~\citep{DBLP:journals/paladyn/RuckstiessSSWSS10,DBLP:journals/nn/SehnkeORGPS10,DBLP:journals/corr/abs-2109-00157}.
One goal of parameter-space exploration has been to find high-quality trajectories from policies that are in the neighborhood of the current policy, thereby mitigating the issue that action-space noise might sample unlikely or low-quality outputs~\citep{DBLP:journals/ftrob/DeisenrothNP13}. Additionally, weight perturbations can directly alter logit values and reorder token distributions as illustrated in \cref{fig: teaser}. Intuitively, if we could learn the amount of noise to add to each parameter, we could explore promising policies that each have promising alterations to token distributions and rollouts. However, to the best of our knowledge, such strategies have been underexplored for LLMs, possibly because noisy optimization itself is seldom used.

In this work, we present \methodfullname{} (\method{}) which uses parameter-space exploration for RLVR by sampling models from an approximate posterior during training that can be updated during the reinforcement learning stage to adaptively learn noise injection.
At each training step, one or more samples are obtained by adding learned noise to the learned parameters, which can be used to control the amount of exploration.
We explore three strategies for noise injection. \batch{} samples a single weight perturbation per gradient step, which makes it dependent on a single weight draw and can limit group diversity. \mcsamples{} calculates advantages separately for $M$ model perturbations before a single weight update. \interleaved{} divides each GRPO group across $N$ independently sampled weights and calculates advantages on the full group to maximize diversity.

We validate \method{} on OLMo-3-1025-7B~\citep{olmo2025olmo3} and Qwen2.5-Math-7B~\citep{yang2024qwen25mathtechnicalreportmathematical} across mathematical reasoning and code generation tasks, which are widely used as RLVR testbeds~\citep{chen2026exploration,DBLP:journals/corr/abs-2505-22617,liu2025understanding,DBLP:journals/corr/abs-2506-01939,zhu2025the}.
Overall, we find that parameter exploration can help downstream performance and speed up convergence without using more rollouts than AdamW-based training. Our chunked noising approach, C3PO, achieves the best average performance across both model families.
Broadly, using multiple model samples per batch performs better than using just a single sample, and rescues more zero-advantage groups than other action-space exploration baselines.
We find consistent improvements in model performance that are concentrated on harder benchmarks like AIME~\citep{aime2026} and LiveCodeBench~\citep{jain2024livecodebench}.
Altogether, our work shows that parameter exploration via learned approximate posteriors can be helpful for RLVR by providing an additional control lever for exploration.

%

\section{Background on RLVR and parameter exploration}
\label{sec: overall_bg}
\subsection{RLVR for LLMs}
\label{sec: rlvr_bg}
Reinforcement Learning with Verifiable Rewards (RLVR) has recently become popular for aligning LLMs in problem settings where outputs of the model for a given problem can easily be verified, for example, in solving mathematical word problems, where a ground-truth solution exists.
Overall, the idea is to assign a reward $r_i$, for example, a binary $r_i \in \{0, 1\}$ to each sequence $\vy^{(i)} \in \vocab^\ast$ that an LLM generates based on a problem $\vx$, which is then subsequently used to train the model.
Crucially, the outputs $\vy^{(i)}$ are self-generated by the model being trained and not obtained via human demonstration. However, some human-provided ground-truth solution is still usually used to calculate $r_i$ on a final answer at the end of the LLM's output $\vy^{(i)}$.

Group Relative Policy Optimization (GRPO)~\citep{DBLP:journals/corr/abs-2402-03300} is a widely-used technique for RLVR training, where the learning signal is calculated based on advantages $A_{i,t} = \frac{r_i - \text{mean}(\text{\vr})}{\text{std}(\text{\vr})}$ using the shorthand $\vr = (r_1, \dots, r_G)$ for a group-size $G$.
That is, a total of $G$ rollouts are sampled for each step that are subsequently used for advantage calculation.
Altogether, the GRPO loss takes the following form, \begin{equation}
    \begin{split}
        \mathcal{L}^{\text{GRPO}} = \mathop{\myexpect}\limits_{\substack{\text{\vx} \sim \data,\; \{\text{\vy}^{(i)}\}_{i=1}^G \sim \pi_{\text{old}}(\cdot|\text{\vx})}} \left[\frac{1}{\sum_{i}|\text{\vy}^{(i)}|} \sum_{i=1}^G \sum_{t=1}^{|\text{\vy}^{(i)}|} \min \left\{R_{i, t} A_{i, t}, \text{clip} (R_{i,t}, 1\!-\!\epsilon, 1\!+\!\epsilon) A_{i,t}\right\} \right].
    \end{split}
    \label{eq: grpo_obj}
\end{equation}
The ratio $R_{i, t}\!=\!\frac{\pi_{\text{$\vparam$}}(y_t^{(i)} \mid \text{$\vx$}, \text{$\vy$}^{(i)}_{<t})}{\pi_{\text{old}}(y_t^{(i)} \mid \text{$\vx$}, \text{$\vy$}^{(i)}_{<t})}$ is an importance sampling factor to account for cases where the language model that generates the rollouts ($\pi_{\text{old}}$) differs from the language model that is updated ($\pi_{\text{$\vparam$}}$). Here, we omit the KL-divergence term following prior work~\citep{yu2025dapo}.

Despite the success of GRPO, several issues remain.
For example, groups that lead to zero-advantages stall learning.
This is the case when all rollouts achieve the same reward and either all lead to a correct or incorrect solution.
Intuitively, this is connected to the problem of trading off exploration and exploitation.
A good RL algorithm needs to mix failed attempts and successful attempts to sustain a learning signal, though achieving a good trade-off is often hard in practice~\citep{DBLP:books/lib/SuttonB98}.
This trade-off is similarly hard to establish in RLVR for LLM training. The entropy of the policy often collapses prematurely, and the groups sampled at each step become increasingly homogeneous, leading to a slowing or even plateauing of learning because the advantages become less informative~\citep{yu2025dapo}.

As a consequence, a variety of approaches have been proposed to target this problem.
Clip-higher~\citep{yu2025dapo} loosens the upper PPO clip to preserve probability mass on exploratory tokens, but it has no effect on on-policy methods as the importance sampling ratio vanishes. Another solution is to increase the softmax temperature during rollout generation in order to flatten the token distribution~\citep{Polaris2025,DBLP:journals/corr/abs-2510-08892}, but this has limited benefits as it is rank-preserving and requires careful tuning to not overshoot. Explicit entropy regularization adds an entropy bonus to the loss~\citep{DBLP:conf/iclr/0003LKD21,DBLP:journals/corr/abs-2509-25133}, and advantage-shaping methods reweight the learning signal toward tokens or trajectories deemed more informative~\citep{DBLP:journals/corr/abs-2506-01939,zhu2025the}. Other strategies are using larger rollout budgets~\citep{DBLP:journals/corr/abs-2510-01180,he2025skywork}, prolonged training~\citep{DBLP:journals/corr/abs-2505-24864,DBLP:journals/corr/abs-2510-01180,DBLP:journals/corr/abs-2510-04028}, or ordering training data by difficulty to gradually expand the reachable solution set~\citep{sun2026rl,DBLP:journals/corr/abs-2506-09026,deepscaler2025}.

Still, all of these methods operate in the action space. They reweigh or regularize token probabilities and do not alter the generators and policy's parameters. In this work, we explore an orthogonal axis: perturbing the policy's parameters directly as an additional axis of exploration in RLVR. By operating in the parameter space, our method is naturally compatible with previous action-space approaches.

\subsection{Parameter space exploration}
\label{sec: param_space_bg}
As previously established, the exploration-exploitation tradeoff is a well-studied problem in classical RL. While methods like $\epsilon$-greedy~\citep{DBLP:conf/nips/Sutton95,watkins1989learning} and UCB~\citep{10.1145/1102351.1102459,DBLP:journals/jmlr/JakschOA10} work well for smaller problems, they tend to not scale well to problems with large action spaces, such as language modeling. Thus, there has been sustained interest in developing efficient algorithms for exploration~\citep{DBLP:journals/corr/abs-2109-00157}.

One such method is parameter exploration, where noise is added to the policy's parameters rather than its actions~\citep{DBLP:journals/paladyn/RuckstiessSSWSS10,DBLP:journals/nn/SehnkeORGPS10}.
Early work used finite-difference gradients with fixed parameter perturbations~\citep{DBLP:conf/icra/KohlS04}; subsequent extensions developed state-dependent and per-basis variants~\citep{DBLP:journals/jmlr/TheodorouBS10}. \citet{DBLP:journals/corr/SalimansHCS17} scaled the approach to deep policies via evolution strategies, and \citet{DBLP:conf/iclr/PlappertHDSC0AA18} and \citet{DBLP:conf/iclr/FortunatoAPMHOG18} integrated parameter noise with modern deep RL algorithms, with the latter learning the noise magnitude jointly with the policy. Across continuous control benchmarks, episode-coherent parameter perturbation has consistently matched or outperformed action-space noise~\citep{DBLP:conf/icml/StulpS12,DBLP:journals/ml/HoofTP17}. Despite these successes, parameter-space exploration methods have largely been absent from RLVR for LLMs. Concurrent work \citep{DBLP:journals/corr/abs-2602-02555} extends the method of \citet{DBLP:conf/iclr/PlappertHDSC0AA18} to RLVR by adding Gaussian noise to the rollout generating policy and all rollouts are generated by the same model. Their approach can be seen as a variant of the \batch{} we present, where the noise added is controlled by an adaptive scheduler rather than a learned Hessian (cf. \cref{eq: sigma}). Additionally, we study the effects of using multiple Monte-Carlo samples (\mcsamples{}) and a chunked noising approach (\interleaved{}) to directly control group diversity in the GRPO objective, which we find is more effective.

\subsection{Variational learning and IVON}
\label{sec: ivon_bg}
In this work, we use variational learning for parameter exploration, because it naturally learns a distribution over neural network parameters.
Given a family of distributions, usually Gaussians, $\mathcal{Q}$, generalized variational learning aims to find the following approximate posterior over parameters $\vparam$, \begin{equation}
    q^\ast(\vparam) = \arg\min_{q\in\calQ} \sum_{i\in \data} \myexpect_{\vparam \sim q} [\ell_i(\vparam) / |\data|] + \lambda^{-1} \dkl{}{q}{p_0}.
    \label{eq: vl}
\end{equation}
The objective is similar to standard training, as it also uses general losses~\citep{KnJe19, KhRu23} but uses a smoothed loss via an expectation over $q$, as well as a KL regularizer towards a prior $p_0$. The prior can be chosen as $p_0 \propto \exp\{-\ell_0\}$, where $\ell_0$ is a standard regularizer like weight decay to mirror conventional deep learning.
The scaling factor $\lambda > 0$ is used to weigh data fit and regularization towards the prior which can be written as an entropy-regularized objective, where $\lambda$ scales the entropy of $q^\ast$ in the objective.

Many methods have been proposed to optimize~\cref{eq: vl}.
For example,~\citet{graves2011practical} and Bayes-by-Backprop~\citep{blundell2015weight} try to use conventional gradient descent methods to learn the mean and variance of a Gaussian $\mathcal{Q}$.
More recently, natural-gradient-based methods have shown more success~\citep{khan2018fast, lin2020handling}.
In particular, the IVON optimizer~\citep{DBLP:conf/icml/ShenDCNMBYGCKM24} has been shown to provide similar performance as methods like AdamW~\citep{DBLP:conf/iclr/LoshchilovH19} at little overhead in terms of runtime that is mainly due to the sampling implementation.

In IVON, at each step the loss is calculated at a perturbed point $\hat{\vparam}$, and not at the current mean $\vm$, which corresponds to the point estimate learned by e.g. AdamW.
Writing it out as the reparameterization, \begin{equation}
    \label{eq: how_to_sample_weights}
    \hat{\vparam} = \mathbf{m} + \mathbf{\sigma} \odot \mathbf{z}; \quad \mathbf{z} \sim \mathcal{N}(0, I),
\end{equation}
shows the similarity to weight-perturbation-based optimization strategies, which have been found to, among others, lead to improved generalization as they avoid sharp minima~\citep{wu2020adversarial, foret2021sharpnessaware, moellenhoff2023sam}.

The diagonal variance in IVON is defined inversely proportional to the Hessian which is weighted by the effective sample size (ESS) $\lambda > 0$ from~\cref{eq: vl}. The ESS can be tuned as an inverse temperature; for example, $\lambda = 10^9$ is common.
Concretely, the variance is defined as follows: \begin{equation}
    \label{eq: sigma}
    \sigma = 1 / \sqrt{\lambda(\vh + \delta)},
\end{equation}
where $\vh$ is the diagonal Hessian initialized to a constant $h_0$, and $\delta \geq 0$ is a weight-decay scaling.
Choosing $\lambda$ large thus reduces noise, while choosing $\lambda$ small increases the amount of noise added to the parameters, which in turn increases diversity in predictions.
We use this strategy to control exploration in reinforcement learning with verifiable rewards next.

\begin{figure}[!t]
    \centering
    \includegraphics[width=\textwidth]{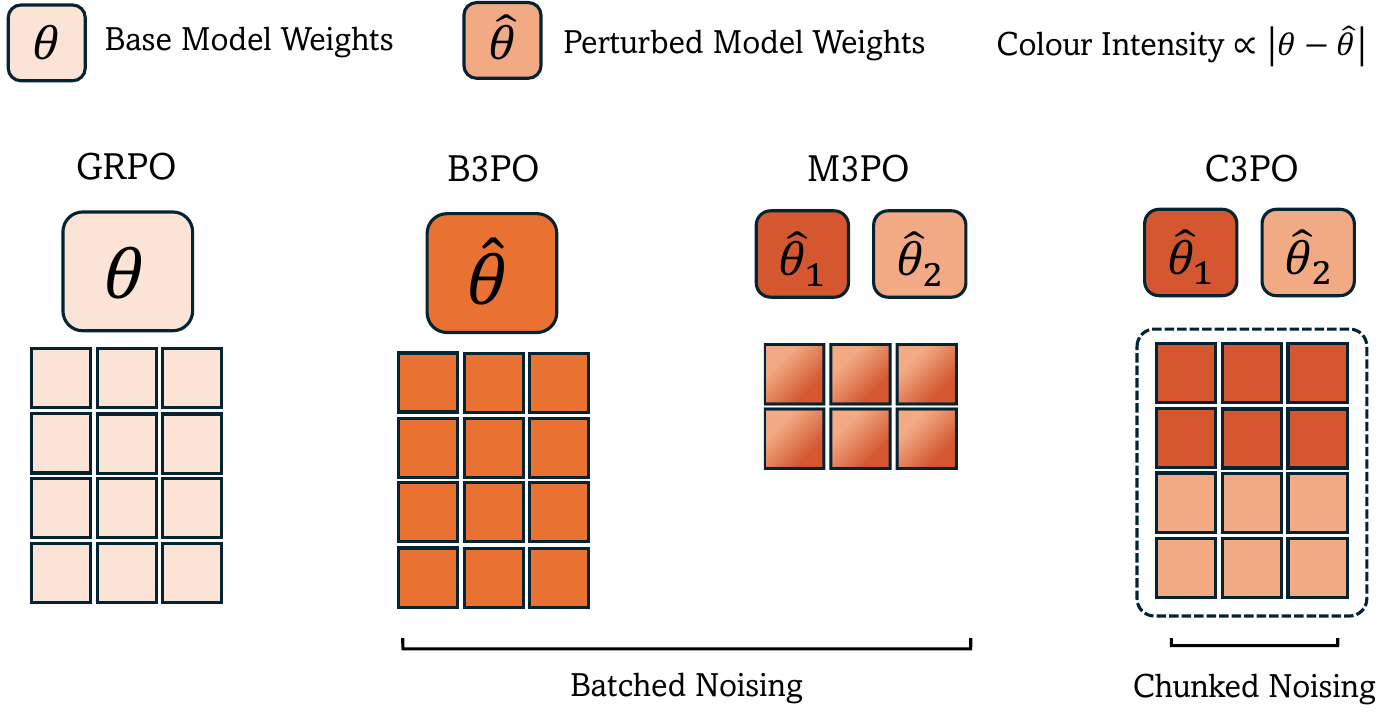}
    \caption{\textbf{Overview of noising strategies.}
        Squircles denote model weights (base $\vparam$ or perturbed $\hat{\vparam}$), while squares denote rollouts; each column corresponds to a rollout group for a different prompt. Color intensity denotes the amount of noise added. \textbf{GRPO} generates all $G$ rollouts using a single shared model $\vparam$, with diversity arising only from sampling. \textbf{B3PO} samples a single perturbation $\smash{\hat{\vparam}}$ per batch and reuses it across all prompts and groups. \textbf{M3PO} accumulates the loss over $M$ perturbations to stabilize training. \textbf{C3PO} partitions rollouts across $N$ perturbations and applies GRPO to the aggregated set, which now contains rollouts from multiple model samples.}
    \label{fig: approach}
\end{figure}


\section{\methodfullname{}}
\label{sec: approach}
We now describe \methodfullname{}, which introduces exploration in the parameter space as an additional axis of exploration in RLVR by using noisy policies. Importantly, our method is compatible with existing approaches like temperature control~\citep{Polaris2025,DBLP:journals/corr/abs-2510-08892}, clip-higher~\citep{yu2025dapo}, and entropy regularization~\citep{DBLP:conf/iclr/0003LKD21,DBLP:journals/corr/abs-2509-25133}. We study three different strategies that are illustrated in \cref{fig: approach}.

\subsection{Batched noising (\batch{} and \mcsamples{})}
\label{sec: batch_level_noise}
Modern RL codebases such as \texttt{verl}~\citep{sheng2024hybridflow} and SkyRL~\citep{cao2025skyrl} decouple generation and gradient updates into separate inference and training engines, periodically synchronizing their weights at the start of every gradient step. Batched noising works well with such implementations, because a single weight perturbation $\smash{\widehat{\vparam}}$ is sampled from the IVON posterior (\cref{eq: how_to_sample_weights}) and synced with the rollout engine. \batch{} is analogous to step-level temperature control~\citep{DBLP:journals/corr/abs-2510-08892}, but acting in the weight space instead, and can be interpreted as a one Monte-Carlo sample approximation of~\cref{eq: vl} using~\cref{eq: grpo_obj} as the loss.

\batch{} could have similar issues as standard GRPO if the single weight perturbation per step does not perform well (or produces homogeneous rollouts).
One way to address this issue is to use multiple MC samples for the expectation in
\cref{eq: vl}, i.e., sampling $M$ perturbations from the IVON posterior $q$. For each, we generate
rollouts over the entire prompt batch and compute the GRPO loss in \cref{eq: grpo_obj}. The posterior is updated once at the end,
with gradients averaged across all $M$ perturbations.
We refer to this strategy as \mcsamples{}. The pseudo-code for both methods is shown in~\cref{alg: batch_level_noise}.

\subsection{Chunked noising (\interleaved{})}
\label{sec: c3po}
GRPO's learning signal stems from within-group reward differences, but M3PO calculates advantages separately for each group instead of using one large group across samples that, intuitively, should be more diverse as it uses rollouts from different models to calculate advantages.
We address this issue in \interleaved{}, where we sample $N$ weight samples using IVON, say with groups of size $G/N$, and then calculate advantages over the full group of $G$ responses, whose diversity now reflects $N$ distinct weight-space samples. The full procedure is given in \cref{alg: interleaved_prompt_level_noise}.

Empirically, we found that Seq-MIS~\citep{liu-li-2025-rl-collapse}, which uses sequence-level importance sampling and masking, was important to stabilize training and ensure convergence (see \cref{sec: seqmis_appendix} for an ablation). A possible reason for this observation is that the policies that are used for gradient estimation differ from the $N$ rollout generators, akin to training\,--\,inference mismatches in RLVR literature~\citep{yao2025offpolicy}. Caching and replaying the noise could be an alternative but comes at increased memory cost.

\section{Experiments}
\label{sec: experiments}

\subsection{Experimental setup}
\label{sec: setup}
We use two base models, namely, \texttt{Olmo-3-1025-7B}~\citep{olmo2025olmo3} and \texttt{Qwen2.5-Math-7B}~\citep{yang2024qwen25mathtechnicalreportmathematical} for our experiments.
We follow a two-stage post-training pipeline, beginning with a warm-start SFT followed by RLVR~\citep{DBLP:journals/corr/abs-2411-15124,deepseekai2025deepseekr1incentivizingreasoningcapability}. For the SFT phase, we use a $\approx\!2$M-example subset of the Llama-Nemotron Post-Training Dataset~\citep{bercovich2025llamanemotronefficientreasoningmodels} containing DeepSeek-R1~\citep{deepseekai2025deepseekr1incentivizingreasoningcapability} responses covering math, code, general reasoning, and instruction following. We train for two epochs using IVON~\citep{DBLP:conf/icml/ShenDCNMBYGCKM24} for both Olmo3 and Qwen2.5-Math. This phase equips the model with initial reasoning capabilities. While it also produces a learned noise distribution that could be reused as an initialization for RLVR, throughout the main results, we initialize the IVON Hessian $\vh$ from scratch with a constant $h_0$ rather than loading the SFT optimizer state to keep our results applicable to any off-the-shelf model and allow us to isolate the benefits of a learned initialization in \cref{sec: learned_vs_scratch}.

Unless specified otherwise, we train our models using GRPO~\citep{DBLP:journals/corr/abs-2402-03300} on DAPO-MATH-17k~\citep{yu2025dapo}, which contains approximately $17,000$ math problems with ground-truth answers. We evaluate on six widely used mathematical reasoning benchmarks: AIME 2024--26, AMC 2023, MATH-500~\citep{DBLP:conf/nips/HendrycksBKABTS21}, and Minerva~\citep{DBLP:conf/nips/LewkowyczADDMRS22}. Across all experiments, we report the mean and standard error for Pass@1, calculated across $K\!=\!8$ rollouts using the unbiased estimator given by~\citet{DBLP:journals/corr/abs-2107-03374}.

We compare \method{} to three action-space exploration baselines: adding an entropy regularization term (EntReg), raising sampling temperature when entropy collapses beyond a threshold (Polaris \citep{Polaris2025}) and applying a KL penalty to tokens with a high covariance between probability and advantage (KL-Cov \citep{DBLP:journals/corr/abs-2505-22617}). Since these methods modify the token distribution while \method{} perturbs the weights, they are complementary rather than competing, and can be applied alongside \method{}.

For all methods, each prompt receives $16$ rollouts per batch. For \mcsamples{}, we report an equal-compute comparison with AdamW, using $M\!=\!4$ MC samples per batch to generate rollouts and compute advantages with $G\!=\!4$ at a time. The resulting policy update is thus calculated on a total of $G\times M\!=\!16$ rollouts. For \interleaved{}, we sample $N\!=\!4$ distinct policies per batch and accumulate $4$ rollouts for each in a rollout buffer, computing advantages on $16$ rollouts in total. We use $\lambda=10^9$ for Olmo3 and $\lambda=10^{10}$ for Qwen2.5-Math. We sweep these hyperparameters in \cref{sec: analysis,sec: lambda_appendix,sec: mc_appendix}, and describe our training configuration in detail in \cref{app: hparams}.

\subsection{Main results}
\label{sec: main_results}
\begin{table*}[!t]
    \centering
    \footnotesize
    \begin{tabular}{llccccccc}
        \toprule
         & \textbf{Method}             & \textbf{AIME '24}               & \textbf{AIME '25}               & \textbf{AIME '26}               & \textbf{MATH-500}               & \textbf{AMC}                    & \textbf{Minerva}                & \textbf{Average}                \\
        \midrule
        \multirow{8}{*}{\rotatebox{90}{\olmologo{}}}
         & \texttt{SFT}                & \valc{16.67}{2.12}              & \valc{21.25}{2.26}              & \valc{12.50}{2.13}              & \valc{78.08}{0.68}              & \valc{51.25}{3.53}              & \valc{30.51}{1.10}              & \valc{35.04}{0.51}              \\
         & \texttt{GRPO}               & \valc{25.42}{2.92}              & \valc{22.08}{2.48}              & \valc{\textbf{19.17}}{3.45}     & \valc{84.93}{0.47}              & \valc{69.38}{3.50}              & \valc{36.99}{0.94}              & \valc{42.99}{0.51}              \\
         & \texttt{GRPO (EntReg)}      & \valc{23.75}{2.35}              & \valc{23.33}{2.45}              & \valc{17.92}{2.68}              & \valc{84.60}{0.58}              & \valc{68.44}{2.89}              & \valc{36.58}{0.95}              & \valc{42.44}{0.53}              \\
         & \texttt{GRPO (Polaris)}     & \valc{26.25}{2.65}              & \valc{20.83}{2.63}              & \valc{15.83}{3.47}              & \valc{84.88}{0.63}              & \valc{64.38}{2.94}              & \valc{36.63}{0.99}              & \valc{41.47}{0.54}              \\
         & \texttt{GRPO (KL-Cov)}      & \valc{24.17}{2.23}              & \valc{24.58}{2.78}              & \valc{16.25}{2.93}              & \valc{83.88}{0.68}              & \valc{66.56}{3.45}              & \valc{\textbf{38.28}}{0.94}     & \valc{42.29}{0.49}              \\
         & \cg \texttt{\batch{}}       & \cg \valc{25.42}{2.76}          & \cg \valc{25.00}{2.34}          & \cg \valc{15.00}{2.34}          & \cg \valc{83.30}{0.58}          & \cg \valc{\textbf{70.31}}{3.01} & \cg \valc{37.36}{0.92}          & \cg \valc{42.73}{0.48}          \\
         & \cg \texttt{\mcsamples{}}   & \cg \valc{25.83}{3.48}          & \cg \valc{24.58}{2.53}          & \cg \valc{17.92}{2.64}          & \cg \valc{\textbf{85.05}}{0.51} & \cg \valc{69.38}{3.27}          & \cg \valc{36.58}{0.83}          & \cg \valc{43.22}{0.49}          \\
         & \cg \texttt{\interleaved{}} & \cg \valc{\textbf{27.50}}{2.91} & \cg \valc{\textbf{26.25}}{2.58} & \cg \valc{\textbf{19.17}}{2.94} & \cg \valc{84.53}{0.48}          & \cg \valc{70.00}{3.43}          & \cg \valc{36.81}{0.93}          & \cg \valc{\textbf{44.04}}{0.46} \\
        \midrule
        \multirow{8}{*}{\qwenlogo{}}
         & \texttt{SFT}                & \valc{14.58}{2.50}              & \valc{20.00}{3.12}              & \valc{12.50}{2.39}              & \valc{65.25}{1.93}              & \valc{59.38}{2.98}              & \valc{24.59}{1.15}              & \valc{32.72}{0.36}              \\
         & \texttt{GRPO}               & \valc{24.17}{3.25}              & \valc{25.42}{3.51}              & \valc{20.42}{3.05}              & \valc{88.20}{0.96}              & \valc{70.63}{3.01}              & \valc{43.38}{1.20}              & \valc{45.37}{0.35}              \\
         & \texttt{GRPO (EntReg)}      & \valc{23.33}{3.34}              & \valc{24.17}{3.67}              & \valc{15.83}{3.19}              & \valc{86.28}{0.99}              & \valc{65.63}{3.06}              & \valc{42.65}{1.18}              & \valc{42.98}{0.34}              \\
         & \texttt{GRPO (Polaris)}     & \valc{22.92}{3.46}              & \valc{26.67}{3.35}              & \valc{21.25}{3.14}              & \valc{87.75}{1.02}              & \valc{72.81}{2.97}              & \valc{43.29}{1.26}              & \valc{45.78}{0.32}              \\
         & \texttt{GRPO (KL-Cov)}      & \valc{22.92}{3.13}              & \valc{24.58}{3.58}              & \valc{20.42}{3.20}              & \valc{87.75}{0.94}              & \valc{71.88}{2.63}              & \valc{43.20}{1.24}              & \valc{45.12}{0.37}              \\
         & \cg \texttt{\batch{}}       & \cg \valc{23.75}{3.14}          & \cg \valc{25.42}{3.65}          & \cg \valc{24.17}{3.19}          & \cg \valc{87.68}{0.96}          & \cg \valc{72.81}{2.62}          & \cg \valc{43.34}{1.21}          & \cg \valc{46.19}{0.37}          \\
         & \cg \texttt{\mcsamples{}}   & \cg \valc{23.75}{3.11}          & \cg \valc{25.42}{3.44}          & \cg \valc{24.17}{3.29}          & \cg \valc{\textbf{88.58}}{0.95} & \cg \valc{\textbf{73.44}}{3.00} & \cg \valc{43.01}{1.20}          & \cg \valc{46.39}{0.36}          \\
         & \cg \texttt{\interleaved{}} & \cg \valc{\textbf{25.00}}{3.21} & \cg \valc{\textbf{27.08}}{3.12} & \cg \valc{\textbf{26.25}}{2.93} & \cg \valc{88.10}{0.93}          & \cg \valc{71.25}{2.67}          & \cg \valc{\textbf{43.61}}{1.22} & \cg \valc{\textbf{46.88}}{0.38} \\
        \bottomrule
    \end{tabular}
    \caption{\textbf{A comparison of all methods across common mathematical reasoning benchmarks}. Methods with multiple noise samples outperform GRPO on average. Both models benefit most from \interleaved{}'s increased group diversity. The largest gains are on the harder AIME problems. We report the unbiased Pass@1 across 8 samples, and its standard error.}
    \label{tab: main_results}
\end{table*}
\textbf{Parameter-space exploration outperforms action-space baselines.} We report benchmark scores across all methods and models in \cref{tab: main_results}. All \method{} variants outperform action-space exploration baselines on average. We hypothesize that these gains are driven by parameter-space exploration discovering fewer invalid rollouts than their action-space counterparts, since trajectories come from different plausible LLM policies. We probe this in \cref{sec: exploration_mechanism}. Notably, performance gains are concentrated on the harder AIME benchmarks, outperforming GRPO by up to $4.17$ points on Olmo3 and $5.83$ points on Qwen2.5-Math. We see similar trends on the relatively harder code generation task in \cref{sec: code_generation}. \interleaved{}'s gain over GRPO is robust to run-to-run variance, measured across 3 seeds (\cref{sec: multiseed_appendix}).

\textbf{On average, using multiple model samples outperforms single-model methods.} \batch{} outperforms GRPO on individual benchmarks such as AIME-2025, AMC, and Minerva, indicating that simply switching optimizers from AdamW to IVON during RLVR could buy modest improvements in some cases. However, its average performance does not improve meaningfully beyond that of the standard GRPO, possibly due to high noise sample variance and limited group diversity. While \mcsamples{} uses $M$ MC samples to reduce noise sample variance and achieves moderate performance gains over \batch{} for both model families, its effectiveness is limited due to a proportional decrease in its group size $G$. We show that lifting the equal-compute constraint further improves the performance of M3PO in \cref{sec: m3po_scaling}. \interleaved{} has the highest average performance in both model families, likely due to the increase in group diversity due to using multiple model samples. We show results with Olmo3 in the next sections and report results using Qwen2.5-Math in \cref{sec: reward_entropy_curves,sec: lambda_appendix,sec: mc_appendix}.

\subsection{Does \method{} really explore more?}
\label{sec: exploration_mechanism}
\begin{figure}[!t]
    \centering
    \includegraphics[width=0.95\textwidth]{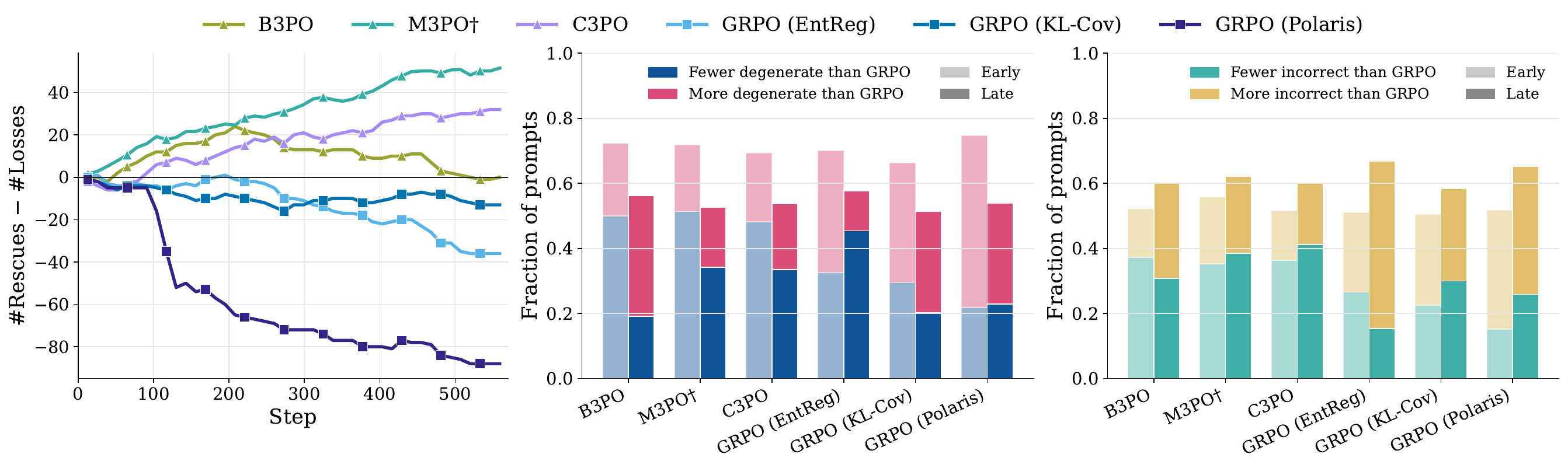}
    \caption{\textbf{(Left) Net zero-advantage groups rescued over training.} \method{} rescues more groups than GRPO and action-space baselines, with multiple perturbations yielding consistent improvements. \textbf{(Middle) Paired degeneracy rates vs GRPO.} Polaris degenerates due to flattening of the token distribution at high temperature. \textbf{(Right) Paired incorrect rollout rates vs GRPO.} While EntReg escapes degeneracy in late training, it buys diversity mostly through incorrect rollouts. Early and late refer to the early (first 33\%) and late training steps (last 33\%), respectively.}
    \label{fig: exploration_mechanism}
\end{figure}

{
    \renewcommand{\thefootnote}{$\dagger$}
    \footnotetext{GRPO's group is subsampled to 4 rollouts for its pairing with M3PO.}
}
In this section, we evaluate whether \method{} can increase exploration during RLVR. However, token-level entropy alone is not sufficient to measure this, since declining entropy could indicate two opposite behaviors. On the one hand, entropy can decrease due to a focus on high-probability\,--\,high-advantage tokens \citep{DBLP:journals/corr/abs-2505-22617} which could be seen as a form of {exploitation}. On the other hand, low entropy can equally signal a policy learning incorrect trajectories~\citep{chen2026exploration}. We observe this phenomenon in \cref{fig: ess_ablation} when noise is scaled too far and where the steep decline in \interleaved{}'s entropy during training (\cref{fig: reward_entropy_curves}) is largely uninformative. Measures such as pass@$k$ may also be misleading, as they measure the ability of a model to explore {post-}RLVR at inference, which is often traded for a higher pass@1 via the entropy\,--\,reward relationship mentioned above in a form of exploitation.\footnote{Still, sampling from the learned posterior could yield inference-time improvements~\citep{bai2025lora}.}

To measure exploration {during} RLVR, we pair each method against GRPO and study how often each method ``rescues'' zero-advantage groups. A prompt is ``rescued'' if a method produces at least one correct rollout where GRPO produced none, and ``lost'' in the opposite case. A successful training-time exploration method should rescue more dead groups than it loses, thus sustaining a learning signal. We also measure the number of degenerate and incorrect rollouts produced by each method during training as compared to GRPO. For each prompt, a method ``wins'' over GRPO if it produces fewer malformed/incorrect rollouts during training. A degenerate (or malformed) rollout is one that does not produce an extractable answer at all, often repeating tokens until the generation limit, and yields no learning signal.

\textbf{\method{} rescues more zero-advantage groups than baselines.} \cref{fig: exploration_mechanism} (Left) shows that \method{} tends to rescue more groups than it loses, while action-space methods fail to do so. Intuitively, 3PO's trajectories come from LLM policies that are likely under the posterior and which are less likely to generate invalid trajectories than the undirected exploration of action-space methods. Using IVON further improves the quality of these trajectories by learning the distribution over policies jointly during training (\cref{eq: sigma}). Polaris loses significantly more groups than it rescues and its gap to GRPO widens during training. Although \batch{} rescues dead groups in early training, its advantage diminishes in the later training steps. Conversely, \mcsamples{} and \interleaved{} continue to rescue zero-advantage groups throughout training, highlighting the benefits of using multiple parameter perturbations.

\textbf{Undirected action-space methods mostly yield malformed or incorrect rollouts.}
Action-space baselines fail to improve over GRPO on malformed and incorrect rollout generation rates. Polaris increases sampling temperature when entropy collapses beyond a threshold. Since temperature only flattens the token distribution (\cref{fig: teaser}), it consistently yields rollouts with no extractable answer (\cref{fig: exploration_mechanism} (Middle)).
EntReg exhibits a different failure mode. Although its rollouts stay well-formed, even beating \method{} in late-training, its higher entropy is spent on mostly incorrect rollouts (\cref{fig: exploration_mechanism} (Right)). KL-Cov produces more degenerate rollouts than GRPO, while its incorrect rate is largely unchanged. \batch{} also tends to produce degenerate rollouts later in training which might explain its diminishing returns in rescuing zero-advantage groups. \mcsamples{} and \interleaved{} are the only methods to consistently produce fewer malformed and incorrect rollouts than GRPO which suggests that parameter-space exploration does produce more useful rollouts for RLVR.

\begin{figure}[!t]
    \centering
    \begin{minipage}[c]{0.59\textwidth}
        \centering
        \includegraphics[width=1.02\linewidth]{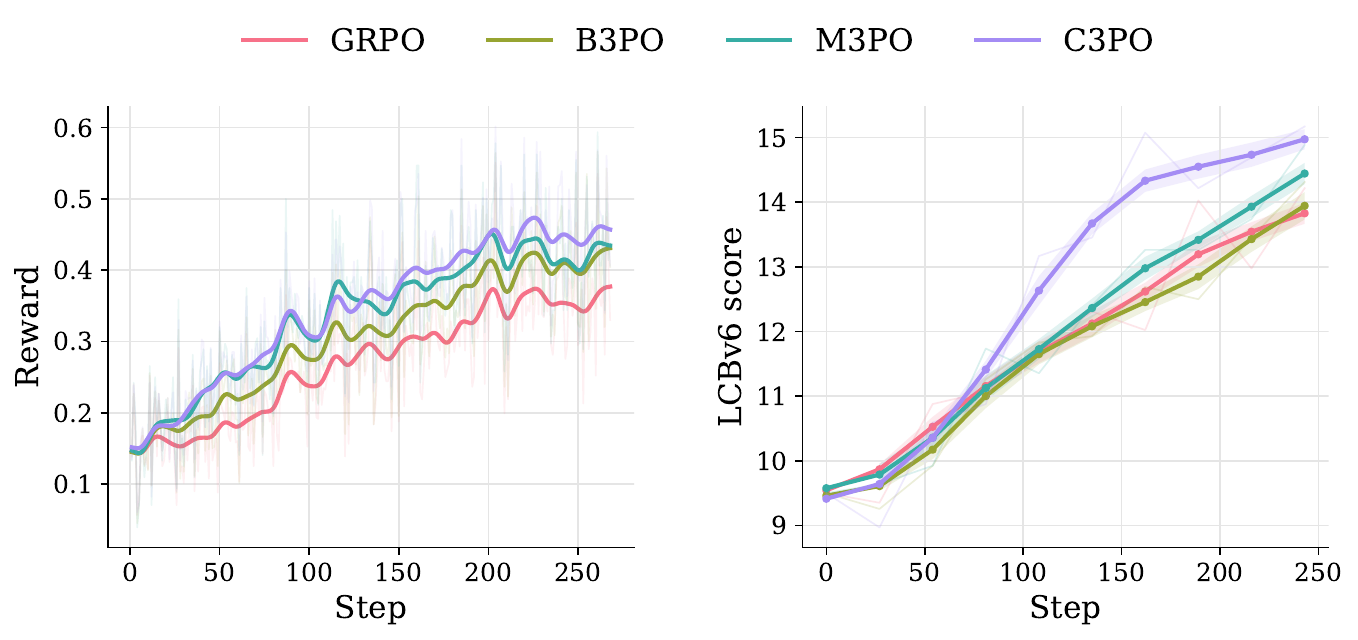}
        \caption{\textbf{Reward and accuracy curves for code generation.} All \method{} methods learn faster and achieve higher rewards than GRPO. \interleaved{} reaches GRPO's final LCB-v6 score at step $162$.}
        \label{fig: codegen_results}

    \end{minipage}
    \hfill
    \begin{minipage}[c]{0.39\textwidth}
        \centering
        \footnotesize
        \begin{tabular}{lcc}
            \toprule
            \textbf{Method}                 & \textbf{ESS ($\lambda$)} & \textbf{Avg. Math Pass@1} \\
            \midrule
            \multirow{3}{*}{\batch{}}       & $10^8$                   & 41.90 $\pm$ 0.51          \\
                                            & $10^9$                   & \textbf{42.73 $\pm$ 0.48} \\
                                            & $10^{10}$                & 42.19 $\pm$ 0.53          \\
            \midrule
            \multirow{3}{*}{\mcsamples{}}   & $10^8$                   & 42.93 $\pm$ 0.51          \\
                                            & $10^9$                   & 43.22 $\pm$ 0.49          \\
                                            & $10^{10}$                & \textbf{43.59 $\pm$ 0.51} \\
            \midrule
            \multirow{3}{*}{\interleaved{}} & $10^8$                   & 0.00 (collapsed)          \\
                                            & $10^9$                   & \textbf{44.04 $\pm$ 0.46} \\
                                            & $10^{10}$                & 41.42 $\pm$ 0.54          \\
            \bottomrule
        \end{tabular}
        \captionof{table}{\textbf{Effects of scaling $\lambda$ for Olmo3:} $10^9$ is a good default value, while $\lambda\!=\!10^8$ hurts performance. \mcsamples{} benefits from a higher $\lambda$.}
        \label{tab: ess_lambda_scaling}
    \end{minipage}
\end{figure}

\subsection{Results for code generation}
\label{sec: code_generation}
In this section, we study the effect of parameter-space exploration beyond mathematical reasoning by testing our methods on code generation tasks. Specifically, given a problem statement, the LLM must produce code that passes all hidden test cases. We train Olmo3 on the CodeR1-12K dataset~\citep{code-r1}, which contains verified coding problems from LeetCode and TACO~\citep{likaixin2024taco-verified}. We use the same hyperparameters as math detailed in \cref{sec: setup} and utilize SandboxFusion~\citep{bytedanceseedfoundationcodeteam2025fullstackbenchevaluatingllms} for executing untrusted LLM code.
We evaluate all models on LiveCodeBench-v6~\citep{jain2024livecodebench}, which contains problems released between January 2025 and April 2025.

As seen in \cref{fig: codegen_results}, there is a consistent gap in the reward curves between our 3PO methods and GRPO. \interleaved{} is the best-performing method in this regime, reaching an LCBv6 score of $15.17$, compared to $13.77$ for GRPO. \interleaved{}'s lead over GRPO emerges around step $100$ and persists for the rest of training, which highlights the effectiveness of increased group diversity through our proposed chunked noising approach. \mcsamples{} narrowly outperforms \batch{} and GRPO, but is probably limited by its lower group diversity due to a proportional decrease in its group size (\cref{sec: m3po_scaling}). Similar to math, both \mcsamples{} and \interleaved{} are more sample efficient and reach GRPO's final reward within the first 50\% of the training steps.

Notably, \method{} methods reach a noticeably higher reward than GRPO on code generation as compared to mathematical reasoning. We hypothesize that this is because code generation is a harder task for the pre-RL model, as seen by the SFT model's performance on the LCBv6 benchmark (9.5\%) compared to the math benchmarks (35.04\%). Thus, increased exploration with \method{} could be more valuable here and aid in discovering high-reward trajectories. We observed a similar trend on the harder AIME benchmarks for math in \cref{sec: main_results}, which suggests that parameter-space exploration is particularly effective for harder tasks where the pre-RL model is less capable and more prone to generating malformed or incorrect rollouts.

\section{Further analysis}
\label{sec: analysis}
We now ablate several components of our proposed approach. Specifically, we study how $\lambda$ should be scaled~(\cref{sec: ess_scaling}), how many MC samples~(\cref{sec: m3po_scaling}) and chunks~(\cref{sec: c3po_scaling}) are needed, and the impact of Hessian initialization (\cref{sec: learned_vs_scratch}). We conclude with practitioner recommendations.
\subsection{Impact of scaling ESS ($\lambda$)}
\label{sec: ess_scaling}

The ESS is a crucial hyperparameter in our setup that controls the amount of noise added to the weights, and needs to be tuned to achieve good performance~\citep{daheim2025uncertaintyaware,bai2025lora}. We ablate $\lambda \in \{10^{8}, 10^{9}, 10^{10}\}$.

We find decreasing $\lambda$ too far to $10^8$ always hurts performance, and completely collapses training for \interleaved{}. This is an expected result because \interleaved{} samples multiple models in the group rollout, but the advantages are still calculated by a single model. Adding too much noise might stray the model too far. Conversely, increasing $\lambda$ to $10^{10}$ has more nuanced results. For the batched noising methods \batch{} and \mcsamples{}, it behaves similarly to $\lambda=10^9$ with a steeper entropy decline than $\lambda=10^8$. $\lambda=10^{10}$ even slightly outperforms $\lambda=10^9$ on average for \mcsamples{} (\cref{tab: ess_lambda_scaling}).

However, $\lambda=10^{10}$ plateaus at a lower ceiling than the $10^9$ \interleaved{} run, as seen from its entropy and reward curves. A possible explanation is that \mcsamples{} has lower variance in the gradient update due to the averaging, and a higher $\lambda$ reduces variance further by reducing the amount of noise added during optimization. On the other hand, \interleaved{} benefits from increased noise due to greater group diversity. We present extended results for this ablation in \cref{sec: lambda_appendix}. It is also plausible that adaptive noise schedules could improve training more. We leave a detailed exploration of this for future work.

\begin{figure}[!t]
    \centering
    \includegraphics[width=0.95\textwidth]{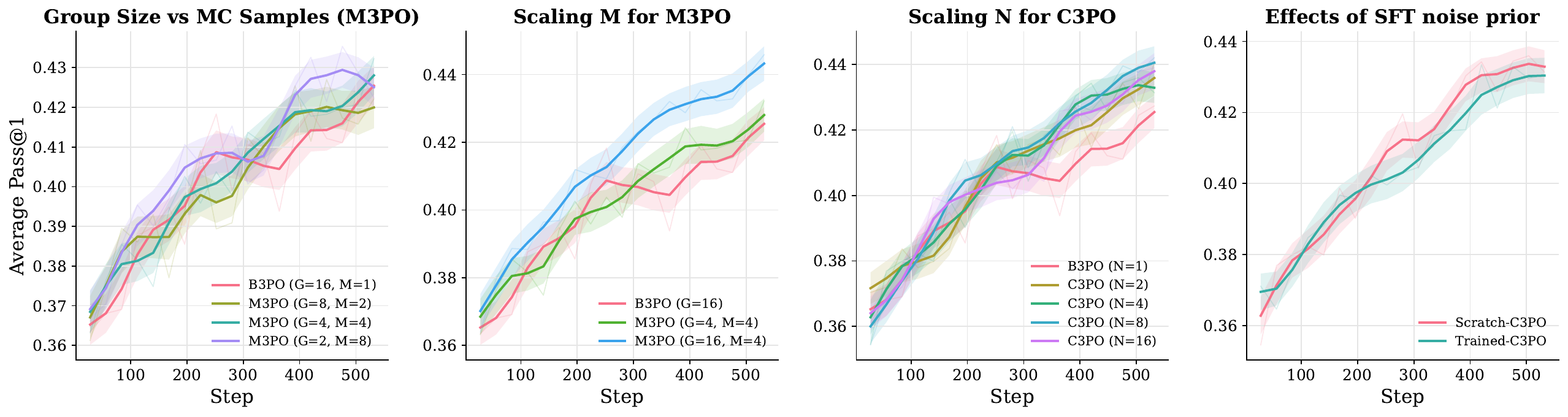}
    \caption{\textbf{(a) Equal-compute comparison:} Benefits of increasing $M$ are offset by reducing $G$. \textbf{(b) Scaling M:} The advantages of more MC samples are fully realized without an equal-compute constraint. \textbf{(c) Scaling N:} $N>1$ improves performance, further scaling is incremental. \textbf{(d) Learned noise prior:} Initializing \vh{} from the learned SFT prior has negligible effects.}
    \label{fig: ablation_grid}
\end{figure}

\subsection{How does the number of MC samples $M$ interplay with group size $G$?}
\label{sec: m3po_scaling}

\mcsamples{} comes at an $M$-fold compute cost because rollouts must be generated $M$ times, once per noise sample (see \cref{tab: wallclock_times}). We offset this by shrinking the per-perturbation group to $G\!=\!16/M$, holding the total rollout count per step fixed. This equal-compute correction introduces an important tradeoff: as $M$ grows, each group shrinks, reducing the diversity in the GRPO group and the chances of sampling a high-reward response per prompt~\citep{DBLP:journals/corr/abs-2510-01180}. \cref{fig: ablation_grid}(a) plots the result of sweeping $M \in \{1, 2, 4, 8\}$ under this constraint. The effects of variance reduction from increasing $M$ are roughly cancelled by the reduced group diversity from a smaller $G$, and vice versa.

Without the equal-compute correction (\cref{fig: ablation_grid}(b)), \mcsamples{} steadily and substantially outperforms the equal-compute run, as well as \batch{} with $G\!=\!16$. This indicates that reduced gradient-update variance does benefit \mcsamples{} when compute is unconstrained, and might explain its advantage over \batch{} in \cref{tab: main_results}. We provide further results in \cref{sec: mc_appendix}.

\subsection{What chunk size $N$ is required?}
\label{sec: c3po_scaling}
We ablate the chunk size $N\!\,\in\!\,\{1, 2, 4, 8, 16\}$ to analyze how many models are required to see the benefit of increased diversity. Note that $N\!=\!1$ reduces to \batch{}.

Even modest group diversification yields a consistent benefit, as all runs with $N \geq 2$ outperform the single-perturbation baseline (\cref{fig: ablation_grid}(c)). Scaling beyond $N=2$ improves performance in earlier training steps but yields smaller gains in final scores. We plot the reward and entropy curves in \cref{sec: chunk_appendix}.

\subsection{Does learning the noise improve performance?}
\label{sec: learned_vs_scratch}
The experiments so far initialize the IVON Hessian $\vh$ to a constant $h_0$ at the start of RLVR, discarding the optimizer state obtained during the warm-start SFT phase. A natural hypothesis is that loading this learned Hessian, which encodes per-parameter variance information from the SFT phase, would yield a better posterior and improve downstream performance. We test this hypothesis by comparing two \interleaved{} runs on Olmo3 that share the same SFT checkpoint. Note that $\vh$ continues to be updated during RLVR in both cases, and replacing it with fixed isotropic noise of matched magnitude drops \interleaved{} below GRPO (\cref{sec: isotropic_appendix}).

\cref{fig: ablation_grid}(d) shows that the two configurations track each other closely throughout training, suggesting that a learned distribution from SFT does not provide a clear advantage. This may be because the SFT phase is relatively short, and the loaded Hessian is nearly isotropic at the end. Concretely, $<\!8\%$ of the values lie outside the $[0.9h_0, 1.1h_0]$ range. Learning the distribution over a longer SFT phase or from pre-training itself could be more beneficial, but we need to leave this experiment for future work due to computational constraints. Nonetheless, our findings transfer to any base model without additional training. We provide detailed results in \cref{sec: learned_prior_appendix}.

\paragraph{Practical recommendations.} Based on our experiments so far, we conclude that using multiple parameter samples per step outperforms a single perturbation. Due to the $G$\!\,--\!\,$M$ tradeoff, we recommend using \interleaved{} with $N\!\,=\,\!2$\,--\,$4$ perturbations and $\vh$ initialized to a constant $h_0$. $\lambda=10^9$ is a good default value, but must be tuned per model. Further hyperparameter details are in \cref{app: hparams}.

\section{Conclusion}
We introduce \methodfullname{} (\method{}), a family of parameter-space exploration strategies for RLVR. By sampling weights from a learned posterior at rollout time, \method{} provides an additional lever for exploration. Across our experiments, drawing multiple model samples per gradient step consistently outperforms action-space baselines. We show that our proposed methods sustain a learning signal by rescuing more zero-advantage groups than action-space baselines. While \mcsamples{} tends to reduce noise sample variance, it either requires more compute or suffers from reduced group diversity. Our chunked noising approach, \interleaved{}, has the best average performance across both model families. We find it useful to tune the amount of added noise through $\lambda$ to control the extent of exploration during RLVR. While a noise prior can be learned before RL, a constant initialization offers similar improvements, making our methods compatible as a drop-in replacement for checkpoints lacking variational training. Overall, our work shows that parameter-space exploration can be used to improve reinforcement learning for large language models.

\section{Limitations}
\label{sec: limitations}
We expect the benefits of \method{} to grow with model scale but our work is limited by compute. \citet{DBLP:journals/corr/abs-2603-12228} show that large LLMs' pretraining neighborhoods are densely populated with task-specific experts. Thus, larger models should tolerate a lower $\lambda$ before training collapses, allowing for more exploration: an exciting direction for future work that we were unable to explore. In addition, \mcsamples{} and \interleaved{} currently have $\approx1.5\times$ higher wall-clock times than GRPO (\cref{app: hparams}), despite equivalent computational requirements in pseudocode. This stems from processing small prompt batches between vLLM model updates, which reduces parallelism benefits~\citep{kwon2023efficient}. Overall, modern RL code stacks all operate using a single model, and it is often difficult to implement changes for variational learning. We also did not explore scaling the number of MC samples and rollouts further for test-time scaling due to limited resources. As with other IVON-based methods, performance is sensitive to the ESS $\lambda$, which must be tuned. Finally, using more directed exploration through structured, non-diagonal posterior estimators~\citep{minut2026soap} instead of IVON, or alternative sampling strategies to Monte-Carlo sampling are a promising direction for future work.


\section*{Acknowledgements and Disclosures}
\paragraph{Funding statement.} We gratefully acknowledge the support from (1) the Ministry of Education and Science of Bulgaria (support for INSAIT, part of the Bulgarian National Roadmap for Research Infrastructure), (2) the hessian.AI Service Center (funded by the Federal Ministry of Research, Technology and Space, BMFTR, grant no. 16IS22091), (3) the hessian.AI Innovation Lab (funded by the Hessian Ministry for Digital Strategy and Innovation, grant no. S-DIW04/0013/003), and (4) the German Federal Ministry of Research, Technology, and Space and the Hessian Ministry of Higher Education, Research, Science, and the Arts within their joint support of the National Research Center for Applied Cybersecurity ATHENE.

\paragraph{Competing interests.} The authors declare no competing interests.

\bibliographystyle{plainnat}
\bibliography{ref}

\appendix
\crefalias{section}{appendix}
\section{Hyperparameter Settings}
\label{app: hparams}
Across our experiments, we use the \texttt{verl} Python library~\citep{sheng2024hybridflow} for training and vLLM~\citep{kwon2023efficient} for efficient inference during rollouts and evaluation. We use an on-policy training setup by performing a single gradient update per prompt batch to avoid performance degradations often associated with batch-online methods~\citep{DBLP:journals/corr/abs-2601-12186,zheng2025groupsequencepolicyoptimization}. Unless otherwise stated, we evaluate with temperature $\tau=0.6$, $\text{top-}p\!=\!0.95$, $\text{top-}k\!=\!50$, and $K\!=\!8$ responses, computing $\text{pass}@k$ with the unbiased estimator of~\citet{DBLP:journals/corr/abs-2107-03374}. We train on an internal cluster of $8$ NVIDIA H200 GPUs with $144$ GB of vRAM, and each algorithm takes $24$\,--\,$40$ hours to run. We report the per-step FLOPs and times in \cref{tab: wallclock_times}. Despite having near-identical FLOPs cost, our approaches that sample multiple models tend to be $\approx\!1.5\times$ slower. This is due to inefficient solutions for sampling from multiple models. Theoretically, the only additional compute over GRPO is the $O(\text{model\_size})$ elementwise noise operations from \cref{eq: how_to_sample_weights} and the recomputation of prefills for each of the $N$ (or $M$) perturbations, both comparatively cheap. As expected, removing the equal-compute constraint on \mcsamples{} ($G\!=\!16, M\!=\!4$) raises the FLOPs cost by a factor of $M$.

For Polaris, we increased the sampling temperature from $1.0$ to $1.4$ when entropy reached $85\%$ of its original value. We experimented with other values like $75\%$ and $50\%$, but these thresholds either triggered the switch too late and did not allow training to recover, or did not trigger at all. We used an entropy coefficient of $\gamma\!=\!10^{-3}$ for EntReg, and additionally experimented with $\gamma\!=\!5\times10^{-3}$ that was too high and completely collapsed training. We use the default hyperparameters swept by the original authors for KL-Cov~\citep{DBLP:journals/corr/abs-2505-22617}.

\begin{wraptable}{r}{0.46\textwidth}
    \centering
    \footnotesize
    \begin{tabular}{l|cc}
        \toprule
        \textbf{Method}                   & \textbf{TFLOPs} & \textbf{Time (mins)} \\
                                          & \textbf{/step}  & \textbf{/step}       \\
        \midrule
        GRPO                              & 51,719          & 2.72                 \\
        \batch{}                          & 51,719          & 2.76                 \\
        \mcsamples{} ($G\!=\!4,M\!=\!4$)  & 51,860          & 4.02                 \\
        \interleaved{} ($N\!=\!4$)        & 52,140          & 4.18                 \\
        \mcsamples{} ($G\!=\!16,M\!=\!4$) & 206,877         & 9.63                 \\
        \bottomrule
    \end{tabular}
    \caption{\textbf{FLOPs and wall clock time per step for each algorithm.} All methods in our main tables have near-identical FLOPs cost. The $1.5\times$ wall-clock gap is a systems artifact of suboptimal multi-model RL infrastructure.}
    \label{tab: wallclock_times}
\end{wraptable}
The $\approx\!1.5\times$ wall-clock overhead is entirely a systems cost stemming from suboptimal RL stacks comprising unfused IVON kernels. The larger cost comes from inefficient multi-model sampling in vLLM. Currently, one must perform $N$ or $M$ inference operations, which reduces the speed gains from vLLM's continuous batching. Implementing more efficient multi-model sampling strategies could significantly reduce this overhead. Moreover, all \method{} variants also converge faster than GRPO (\cref{fig: reward_entropy_curves,fig: codegen_results}), so their effective cost-to-target could even fall below GRPO once these implementations mature. \method{} also introduces no additional memory complexity over AdamW since IVON's optimizer state consists of a momentum and a diagonal Hessian, occupying the same footprint as AdamW's two moment estimates.

\paragraph{Warm-start SFT phase.} We use the IVON optimizer with a learning rate of $50.0$, weight decay $10^{-8}$, $\lambda=10^{10}$, $\beta_1=0.9$, $\beta_2=0.9999$, $h_0=0.001$, and clipping radius $0.001$. We filter the Llama-Nemotron Post-Training Dataset~\citep{bercovich2025llamanemotronefficientreasoningmodels} to retain DeepSeek-R1 responses with a context length of up to $4096$ tokens. We use a cosine-decay learning rate schedule with a $10\%$ linear warmup, decaying to $10\%$ of the initial value by the end of training.

\paragraph{RLVR phase.} We use a token-level GRPO formulation~\citep{yu2025dapo}, with the lower and upper PPO importance-sampling clip bounds both set to $0.2$. We use a binary correctness reward $(+1/0)$, a batch size of $32$ prompts with $16$ rollouts each, and one gradient step per rollout batch. For the AdamW runs, we use a learning rate of $10^{-6}$, weight decay $0.1$, $\beta_1=0.9$, and $\beta_2=0.999$. For IVON, we retain the same hyperparameter configuration as in the warm-start SFT phase, except with a learning rate of $1.0$ and $\lambda=10^9$ for Olmo3 and $\lambda=10^{10}$ for Qwen2.5-Math. We mask out the sequence-level importance weights that lie outside $[0.5, 2.0]$ for \interleaved{}. All runs use a constant learning rate schedule with a $6\%$ linear warmup. Rollouts are sampled at a temperature of $1.0$ during training, and allowed to generate a maximum of $4096$ tokens for $559$ steps for mathematical reasoning and $270$ steps for code generation.

\section{Reward and entropy curves}
\label{sec: reward_entropy_curves}
\begin{figure}[!t]
    \centering
    \includegraphics[width=0.95\textwidth]{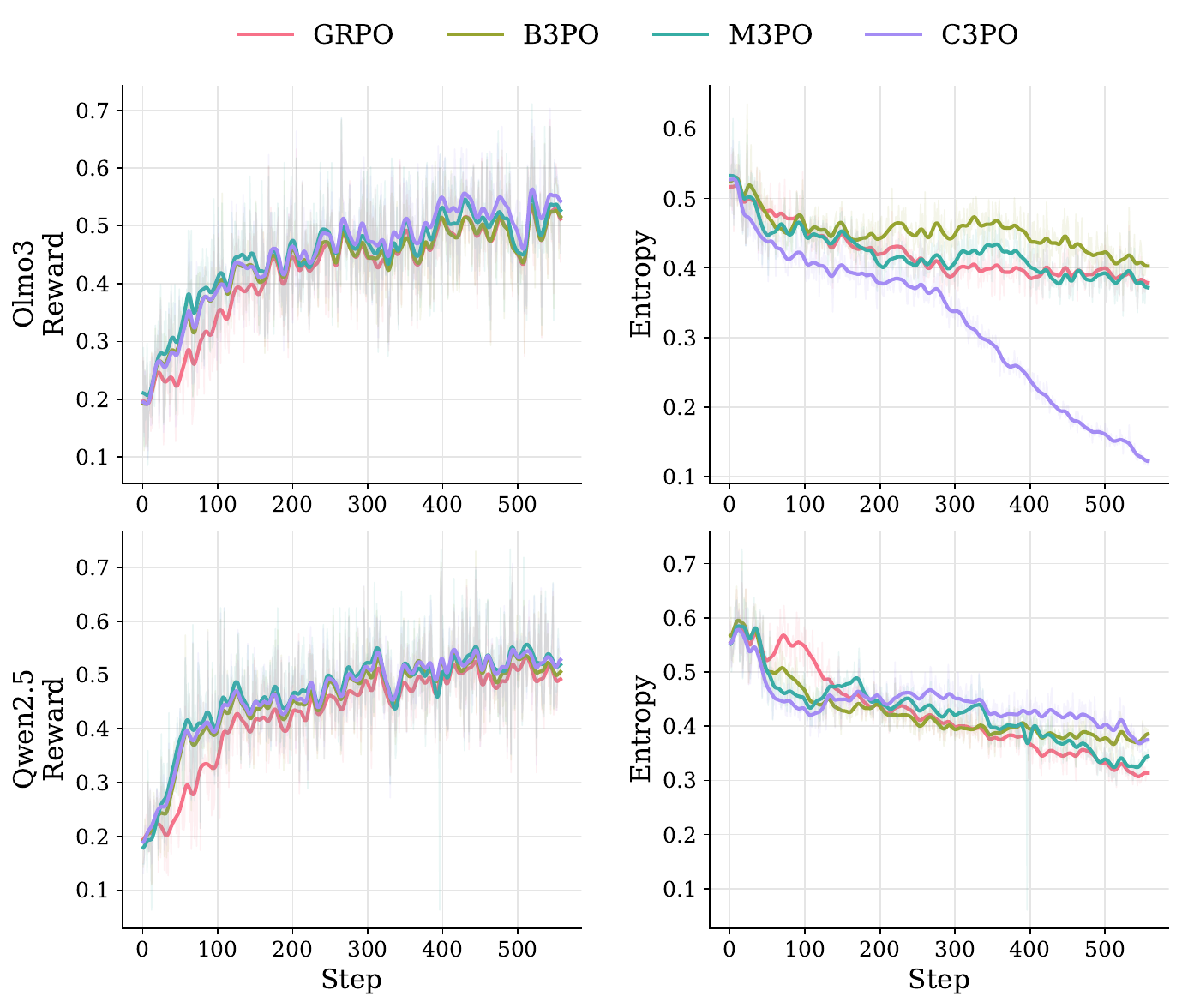}
    \caption{\textbf{Reward and entropy curves for math}. All \method{} methods converge faster than GRPO. Olmo3-C3PO's entropy declines rapidly, but all methods have similar profiles for Qwen2.5-Math.}
    \label{fig: reward_entropy_curves}
\end{figure}

We plot the reward and entropy curves for our proposed methods and GRPO across both model families in \cref{fig: reward_entropy_curves}. Similar to the observations for code generation, our proposed methods consistently converge faster than GRPO across both model families. For Olmo3, \interleaved{}'s entropy declines rapidly, suggesting that it is able to concentrate probability mass on high-reward regions of the solution space. For Qwen2.5-Math, all methods maintain largely similar entropy profiles, indicating that the improved exploration is not simply due to increased randomness in the policy. As mentioned in the main text, we advise against using entropy as a metric for exploration, as it is not always indicative of downstream performance.

\section{Impact of scaling ESS ($\lambda$)}
\label{sec: lambda_appendix}
In the main text, we discussed the importance of $\lambda$ as a crucial control knob for the amount of noise added to the weights, and briefly ablated its effect on our methods for Olmo3. Here, we present detailed results from this ablation.
\begin{figure}[!t]
    \centering
    \includegraphics[width=\textwidth]{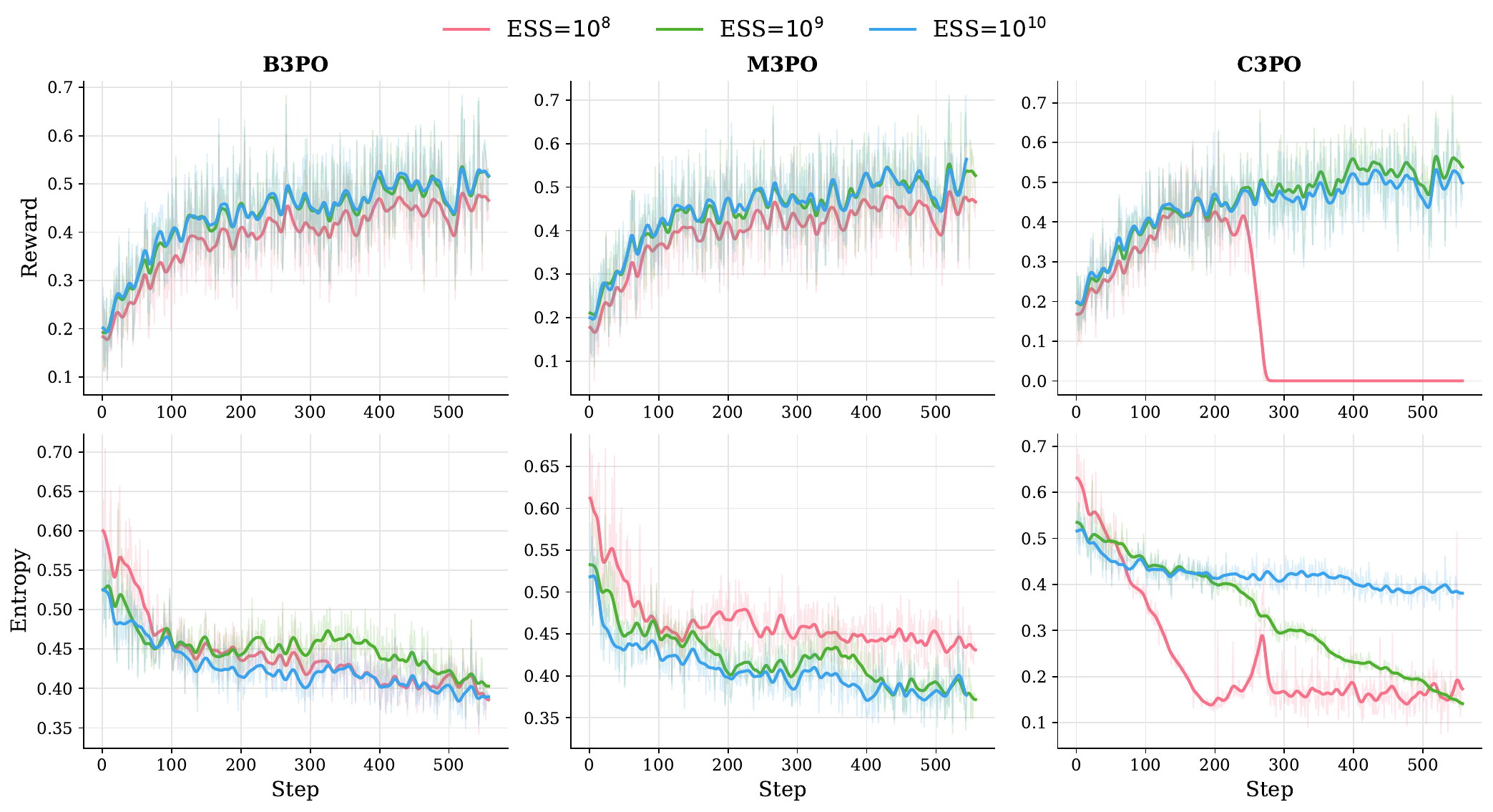}
    \caption{\textbf{$\lambda$ scaling curves across all \method{} methods.} Adding too much noise with small $\lambda$ hurts performance, while large $\lambda$ values can make the sampled models too similar.}
    \label{fig: ess_ablation}
\end{figure}

As noted in \cref{sec: ess_scaling}, $\lambda=10^9$ is a good default value for Olmo3, yielding consistent gains for all three methods. $\lambda=10^8$ tends to be very unstable, generally underperforming higher values and completely collapsing for C3PO. The entropy curve for C3PO at $\lambda=10^8$ is evidence for why entropy curves alone are not reliable indicators of exploration.

\begin{figure}[!t]
    \centering
    \includegraphics[width=0.80\textwidth]{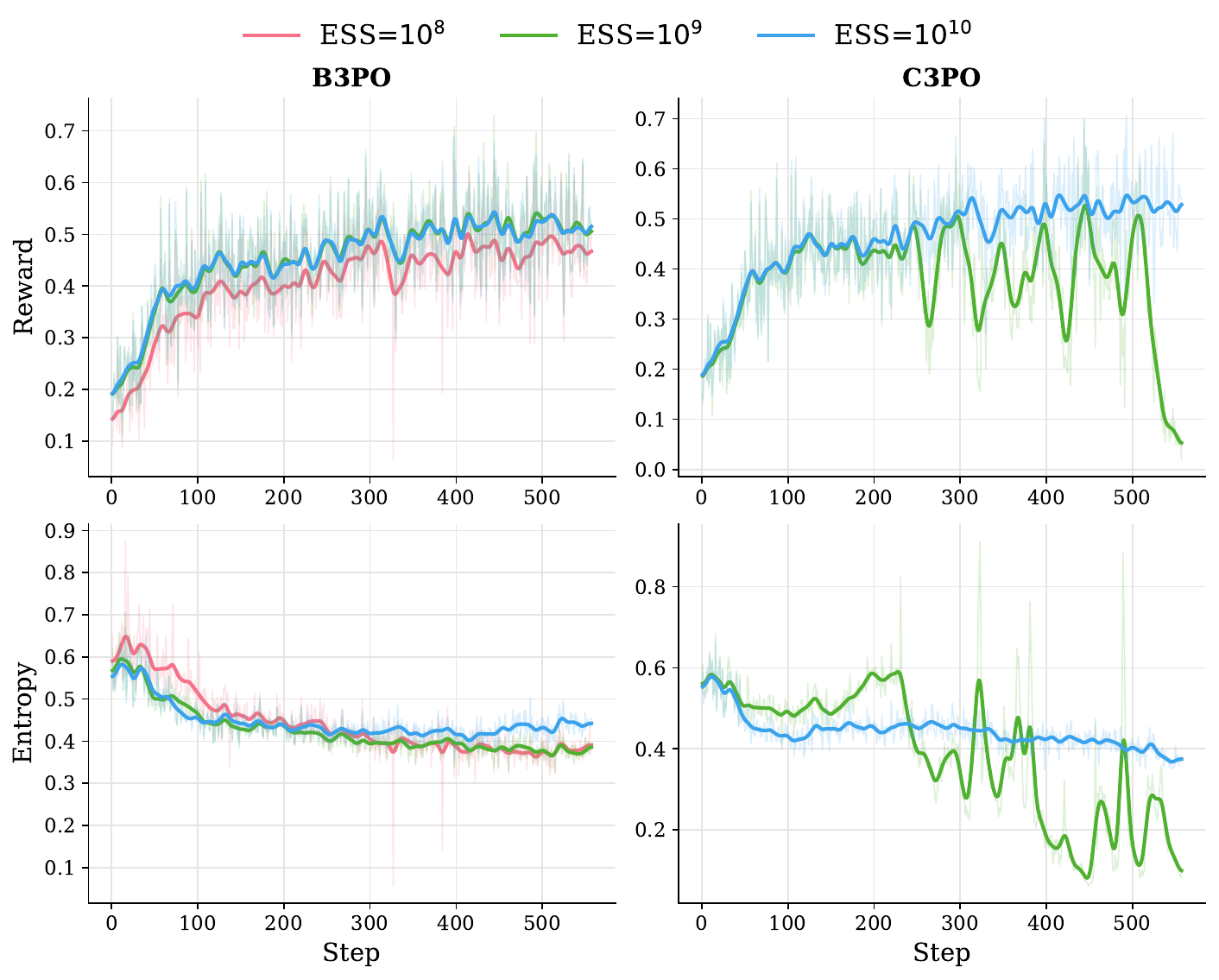}
    \caption{\textbf{Effects of scaling $\lambda$ for Qwen2.5-Math.} Qwen2.5-Math is more sensitive to small $\lambda$, where overexploration causes the curves to oscillate. Tuning $\lambda$ is therefore important.}
    \label{fig: qwen_ess_curve}
\end{figure}

We additionally reproduce the same comparison on B3PO and C3PO for Qwen2.5-Math in \cref{fig: qwen_ess_curve}. Qwen2.5-Math is less robust to small $\lambda$ than Olmo3: while \batch{} behaves similarly across the two models, \interleaved{} is more sensitive to $\lambda$ because it samples a fresh model many more times during rollout generation, producing very erratic reward and entropy curves at low $\lambda$. The default $\lambda=10^{9}$ is too noisy for \interleaved{} with Qwen2.5-Math, and $10^{10}$ works better, which we use in the main text.

\section{Ablating the number of Monte Carlo samples in \mcsamples{}}
\label{sec: mc_appendix}
\begin{figure}[!t]
    \centering
    \includegraphics[width=0.95\textwidth]{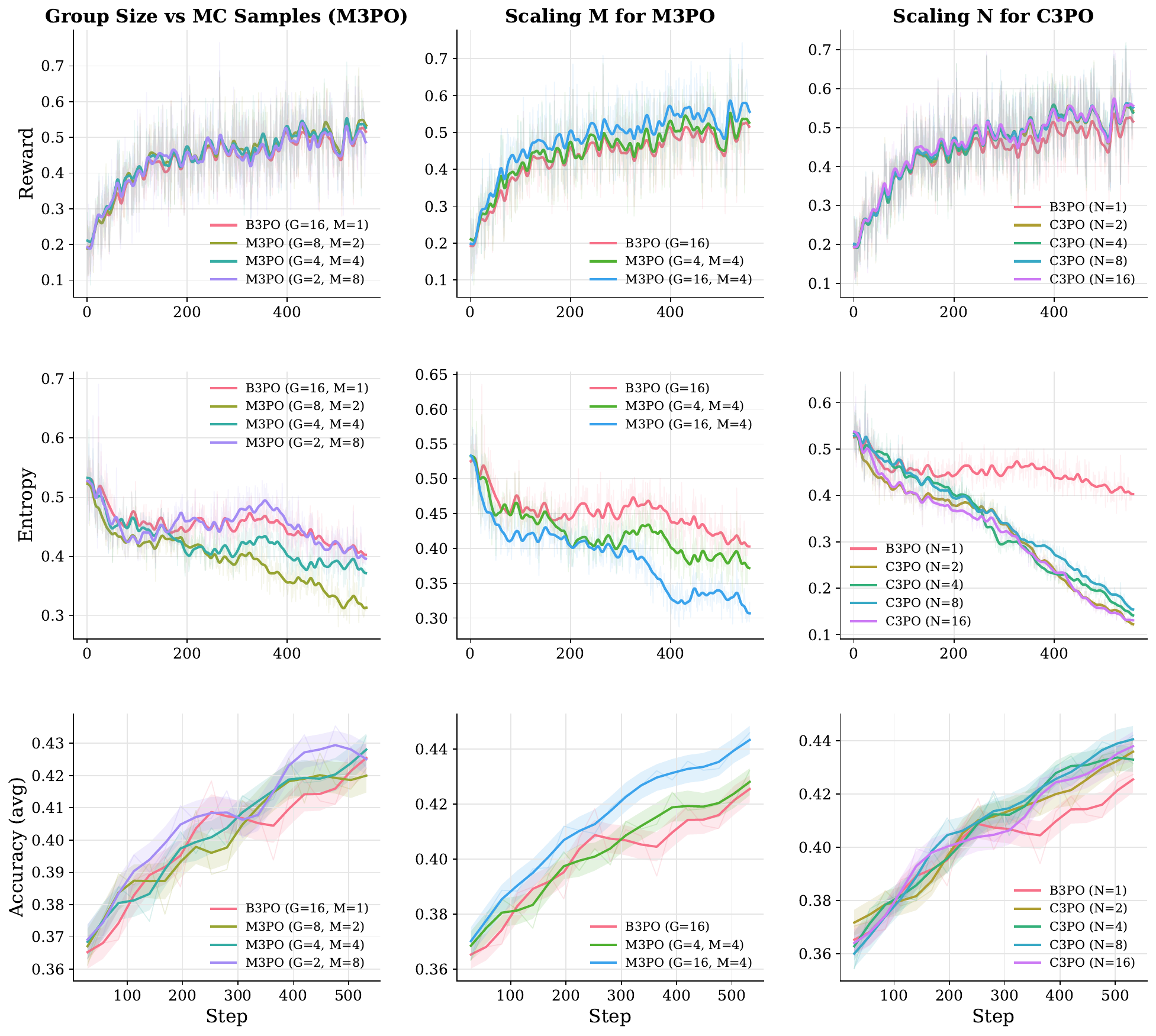}
    \caption{\textbf{Detailed results for ablating MC samples and chunk size.} Using $M>1$ at a constant rollout budget lowers entropy, but average reward and downstream pass@1 are largely unchanged.}
    \label{fig: detailed_ablation_grid}
\end{figure}

\begin{figure}[!t]
    \centering
    \includegraphics[width=0.95\textwidth]{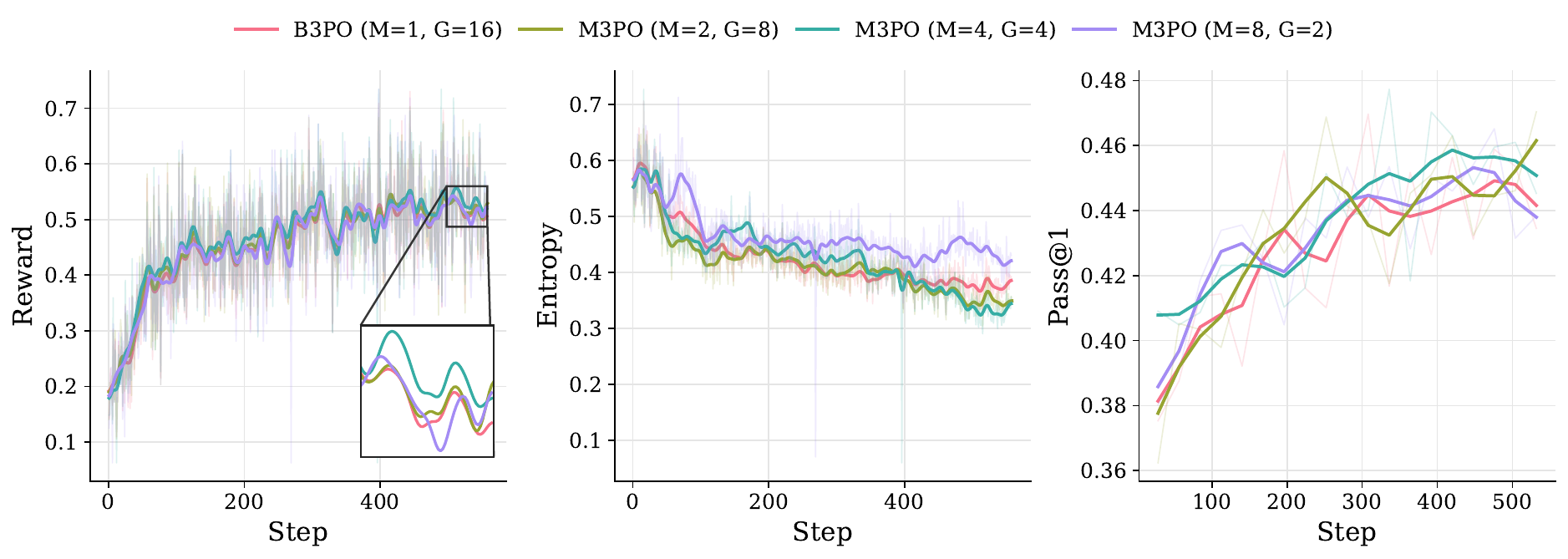}
    \caption{\textbf{$G$--$M$ tradeoff on Qwen2.5-Math.} The variance reduction from larger $M$ does not compensate for the loss in group diversity from smaller $G$, and vice versa.}
    \label{fig: qwen_mcsamples}
\end{figure}

In \cref{sec: m3po_scaling}, we analyzed the tradeoff between the group size $G$ and the number of MC samples $M$ for \mcsamples{}. Increasing $M$ reduces gradient variance but proportionally increases compute, so we shrink $G$ to keep the rollout budget fixed. However, smaller $G$ also reduces group diversity, which is crucial for grouped-advantage algorithms like GRPO~\citep{he2025skywork,DBLP:journals/corr/abs-2510-01180}. We presented the pass@1 scores in \cref{fig: ablation_grid}(a-b); here we additionally plot the corresponding reward and entropy curves in \cref{fig: detailed_ablation_grid}(a-b). Moving from $M=1$ to $M=2$ markedly lowers the entropy curve, but the reward curve plateaus only slightly higher and average downstream performance is unaffected. Scaling $M$ further (at the cost of $G$) progressively closes the entropy gap. However, removing the equal-compute constraint substantially boosts \mcsamples{}, lifting reward and trading more entropy for downstream performance. This suggests that scaling $M$ without shrinking the rollout group could be a viable strategy for improving performance, but it comes with a proportional computational cost.

We also study the $G$--$M$ tradeoff for \mcsamples{} on Qwen2.5-Math. Following \cref{sec: m3po_scaling}, we hold the total rollout count fixed, varying $M\in\{1,2,4,8\}$ and reducing the group size proportionally. \cref{fig: qwen_mcsamples} shows the same trend as on Olmo3: the gain from variance reduction is roughly cancelled by the loss in group diversity. As discussed in \cref{sec: m3po_scaling}, realizing the variance-reduction benefit requires investing additional compute.

\section{Scaling the chunk size in \interleaved{}}
\label{sec: chunk_appendix}
In this section, we examine the effect of varying the chunk size $N$ in \interleaved{} in more detail. \cref{fig: detailed_ablation_grid}(c) shows the reward and entropy curves corresponding to the analysis in \cref{sec: c3po_scaling}, and the trends mirror those of the pass@1 curves. While larger $N$ yields marginally higher reward, even $N=2$ captures almost all of the performance gains. The entropy trend matches \cref{fig: reward_entropy_curves}: $N>1$ sharply increases group diversity, enabling the policy to keep discovering high-reward regions of the solution space and concentrate probability mass there, trading entropy for downstream performance~\citep{DBLP:journals/corr/abs-2505-22617, DBLP:journals/corr/abs-2510-04028}. Together, these results suggest that setting $N$ between $2$ and $4$ and relying on temperature sampling for the remaining rollouts is an effective strategy for improving downstream performance.

\begin{figure}
    \centering
    \includegraphics[width=0.95\textwidth]{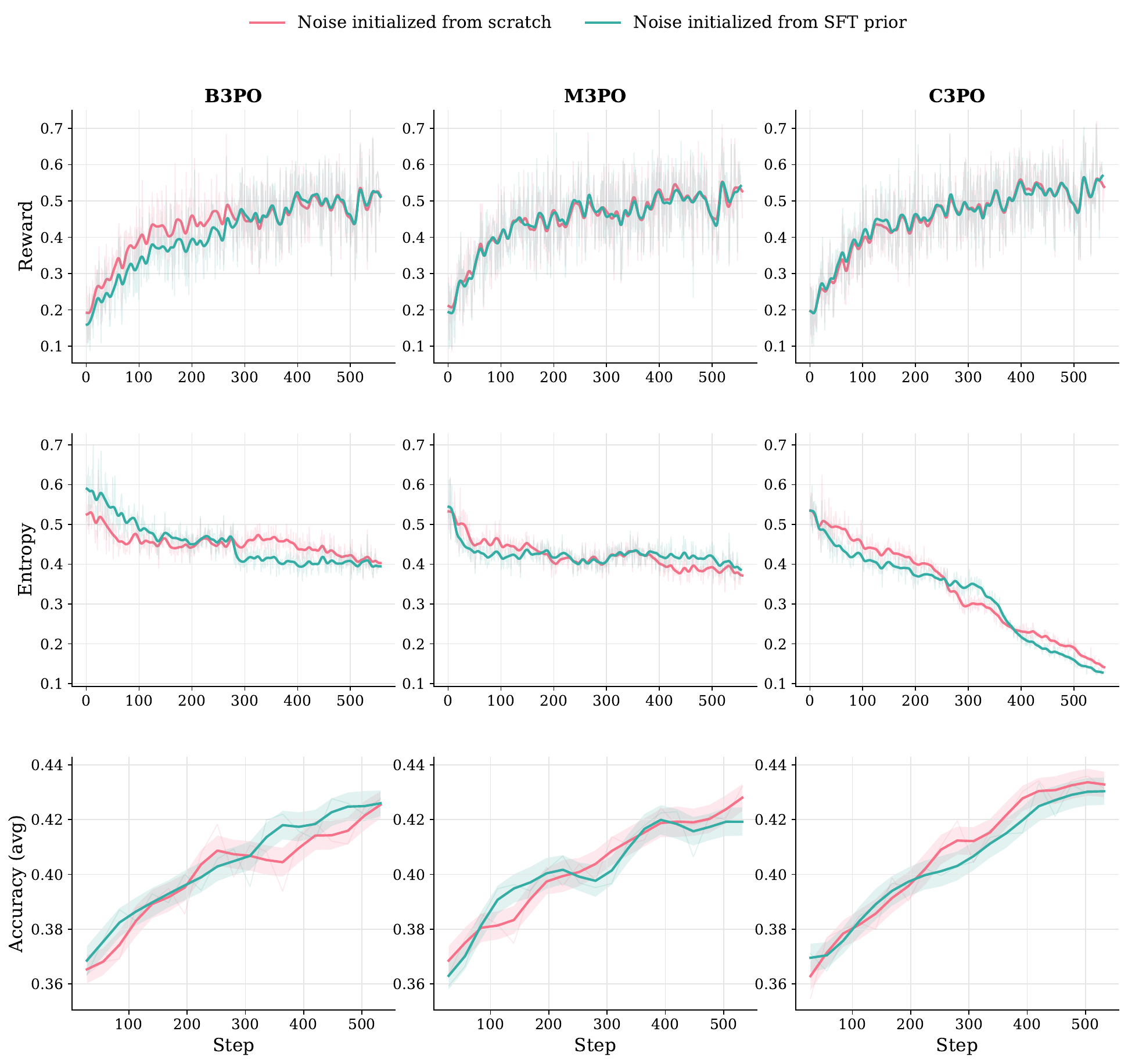}
    \caption{\textbf{Effects of a learned noise prior on other algorithms.} A learned noise prior gives no clear benefit for any \method{} variant and slows \batch{}'s convergence, possibly because the Hessian remains nearly isotropic after SFT.}
    \label{fig: detailed_scratch_vs_trained}
\end{figure}

\section{Effects of a learned noise prior on other algorithms}
\label{sec: learned_prior_appendix}
In \cref{sec: learned_vs_scratch}, we analyzed the impact of initializing the Hessian $\vh$ with a learned prior obtained during the warm-start SFT phase for \interleaved{}. We present detailed results for this comparison across the other 3PO methods in \cref{fig: detailed_scratch_vs_trained}. Intuitively, a learned prior should stabilize learning more than initializing from scratch and thereby improve performance. We do not, however, observe this behavior for any of our methods; for \batch{}, the learned prior even slows convergence. We observed a similar pattern when using Qwen2.5-Math as well. A likely reason is that our SFT phase is relatively short, leaving the Hessian largely isotropic at the start of RL; learning the Hessian from pretraining itself is a promising direction for future work. In the meantime, the absence of a measurable advantage from the learned prior implies that \method{} can be applied to any off-the-shelf checkpoint without first running an IVON-based SFT phase to calibrate the noise distribution.

\begin{figure}[!t]
    \centering
    \includegraphics[width=0.95\textwidth]{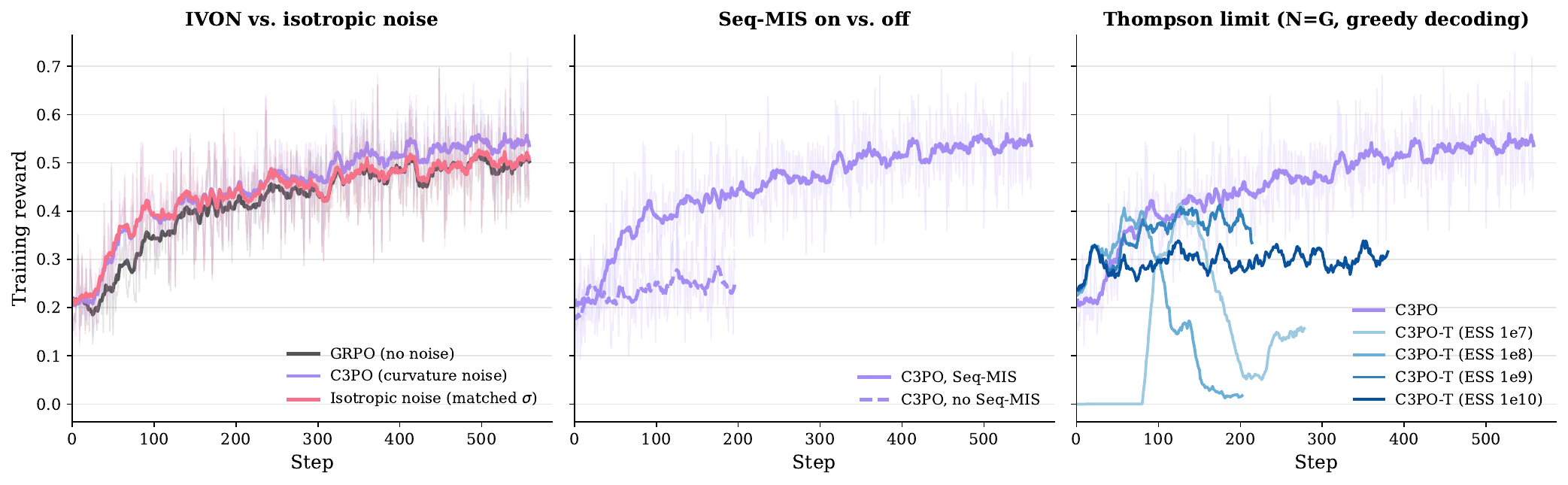}
    \caption{\textbf{(Left) IVON vs.\ fixed isotropic noise.} Isotropic noise climbs faster early but plateaus at a level similar to GRPO. \textbf{(Middle) Effect of Seq-MIS correction.} Training completely stalls without the correction due to training-inference mismatch. \textbf{(Right) Effect of Thompson sampling.} Sampling a fresh model for each rollout and acting greedily under it is unstable.}
    \label{fig: c3po_ablations}
\end{figure}

\section{Multi-seed robustness}
\label{sec: multiseed_appendix}
Here, we validate the robustness of our methods to different seeds. However, because rerunning each method multiple times was not possible due to computational load we only compare GRPO and \interleaved{} on Olmo3 across three seeds on mathematical reasoning tasks. \interleaved{} improves over GRPO by an average of $+0.89$ points across the six benchmarks, positive on all three seeds, with a paired-$t$ $p=0.027$. The gains concentrate on harder benchmarks, with AIME'24 improving by $+1.8$ points ($p=0.023$). These improvements therefore survive run-to-run variance, and combined with the code-generation results (\cref{sec: code_generation}) reinforce that our parameter-space exploration methods have the greatest benefits for harder tasks.

\section{Fixed isotropic noise vs. learned curvature}
\label{sec: isotropic_appendix}
To isolate the contribution of the Hessian-scaled noise (\cref{eq: sigma}), we run \interleaved{} with isotropic noise of matched magnitude, replacing per-parameter variance with a single global scale. This variant attains an average accuracy of $42.12$, below GRPO's $42.99$ and well below full \interleaved{}. As seen in \cref{fig: c3po_ablations} (Left), its reward does climb faster than GRPO early in training but converges to a similar plateau, whereas full \interleaved{} plateaus above GRPO in late training.

\section{Ablating the Seq-MIS correction for \interleaved{}}
\label{sec: seqmis_appendix}
\interleaved{} generates the rollouts of a group from $N$ distinct perturbed models, which introduces a training\,--\,inference mismatch relative to the single policy assumed by the GRPO ratio in \cref{eq: grpo_obj}. To show that correcting this mismatch is necessary, we ran an early \interleaved{} configuration that did not apply any such correction (\cref{fig: c3po_ablations} (Middle)). This run's training reward curve stayed essentially flat throughout, in contrast to the steadily rising reward of our corrected runs. This mirrors the instability that the broader literature attributes to the training\,--\,inference mismatch in RLVR~\citep{yao2025offpolicy,liu-li-2025-rl-collapse}, and motivates the correction used in all of our main results.

\section{A limiting case of \interleaved{}}
\label{sec: thompson_appendix}
One limiting case of \interleaved{} is to use $N\!=\!G$ with temperature $\tau\!=\!0.0$, which would draw a fresh model for each rollout and decode greedily under it. However, we found this variant was very unstable (\cref{fig: c3po_ablations} (Right)), with the reward curve oscillating wildly. Our most stable run at $\lambda=10^{10}$ performed markedly worse than \interleaved{} with temperature sampling. Thus, our methods in the main text perform exploration in both the action and parameter spaces: by first sampling a policy from the IVON posterior and further sampling actions under this policy.

\newpage
\section{Algorithms}
\label{app: algorithms}
\begin{center}
    \begin{tcolorbox}[
            colframe=gray,
            colback=white,
            boxrule=0.5pt,
            arc=2pt,
            left=4pt, right=4pt, top=4pt, bottom=4pt,
            width=0.98\linewidth,
            boxsep=1pt,
            enhanced
        ]
        \begin{algorithmic}[1]
            \Require Dataset $\mathcal{D}$, train policy $\pi_\theta$, rollout policy ${\pi_\mu}$
            \Constants group size $G$, temperature $\tau$, Monte-Carlo samples $M$
            \State $(\mathbf{m},\,\boldsymbol{\sigma}) \gets \operatorname{IVON}(\theta_0)$
            \For{each batch $\mathcal{B} \sim \mathcal{D}$}
            \State $\mathcal{L} \gets 0$
            \For{$j = 1, \dots, M$} \Comment{Optional. $M=1$ for \batch{}}
            \State Sample $\hat{\theta}_j \gets \mathbf{m} + \boldsymbol{\sigma} \odot \mathbf{z}$,\enskip $\mathbf{z} \sim \mathcal{N}(0,\,I)$

            \State $\pi_\mu,\,\pi_\theta \gets \hat{\theta}_j$ \Comment{sync rollout and train policy}
            \For{each $x \in \mathcal{B}$}:
            \State sample $\{o_i\}_{i=1}^{G} \sim \pi_\mu(\cdot \mid x, \tau)$ \Comment{generate $G$ completions per prompt}
            \EndFor

            \State Compute rewards $r_i$ and advantages $A_{i,t}$

            \State $\mathcal{L} \gets \mathcal{L} + \mathcal{L}_{\mathrm{GRPO}}\!\left(\pi_\theta, A_{i,t}\right)$ \hfill (cf.\ Eq.~\ref{eq: grpo_obj})

            \EndFor
            \State $\mathbf{m}, \vsigma \gets \operatorname{IVON.step}(\nabla_\theta \mathcal{L})$ \Comment{Update posterior}
            \EndFor
        \end{algorithmic}
    \end{tcolorbox}
    \vspace{-0.5em}
    \captionof{alg}{\textbf{Batched noising methods (\batch{} and \mcsamples{}).} Weight perturbations $\hat{\theta}$ are sampled once per gradient step; all $G$ rollouts for a batch of prompts are generated from the same model $\pi_\mu$. Model weights are updated by accumulating losses over $M$ Monte Carlo samples; $M\!=\!1$ for \batch{}.}
    \label{alg: batch_level_noise}
\end{center}

\begin{center}
    \begin{tcolorbox}[
            colframe=gray,
            colback=white,
            boxrule=0.5pt,
            arc=2pt,
            left=4pt, right=4pt, top=4pt, bottom=4pt,
            width=0.98\linewidth,
            boxsep=1pt,
            enhanced
        ]
        \begin{algorithmic}[1]
            \Require Dataset $\mathcal{D}$, train policy $\pi_\theta$, rollout policy $\pi_\mu$
            \Constants group size $G$, temperature $\tau$, chunk size $N$

            \State $(\mathbf{m},\,\boldsymbol{\sigma}) \gets \operatorname{IVON}(\theta_0)$

            \For{each batch $\mathcal{B} \sim \mathcal{D}$}

            \State $\mathcal{R} \gets \emptyset$ \Comment{initialize response buffer}

            \For{$n = 1, \ldots, N$}

            \State Sample $\hat{\theta}_n \gets \mathbf{m} + \boldsymbol{\sigma} \odot \mathbf{z}$,\enskip $\mathbf{z} \sim \mathcal{N}(0,\,I)$

            \State $\pi_\mu \gets \hat{\theta}_n$ \Comment{sync rollout policy}

            \For{each $x \in \mathcal{B}$}:
            \State Sample $\bigl\{o_i^{(n)}\bigr\}_{i=1}^{G/N} \sim \pi_\mu(\cdot \mid x, \tau)$;
            \State $\mathcal{R} \gets \mathcal{R} \cup \bigl\{o_i^{(n)}\bigr\}$
            \EndFor

            \EndFor \Comment{$\mathcal{R}$ accumulates $G$ rollouts per prompt from $N$ perturbations}

            \State Compute rewards $r_i$ and advantages $A_{i,t}$

            \State Sample $\hat{\theta}' \gets \mathbf{m} + \boldsymbol{\sigma} \odot \mathbf{z}'$,\enskip $\mathbf{z}' \sim \mathcal{N}(0,\,I)$;\enskip $\pi_\theta \gets \hat{\theta}'$ \Comment{fresh draw for the update}

            \State $\rho_i^{(n)} \gets \pi_\theta\bigl(o_i^{(n)} \mid x\bigr) / \pi_{\hat{\theta}_n}\bigl(o_i^{(n)} \mid x\bigr)$; mask $o_i^{(n)}$ if $\rho_i^{(n)} \notin [0.5, 2]$ \Comment{Seq-MIS}

            \State $\mathcal{L} \gets \mathcal{L}_{\mathrm{GRPO}}\!\left(\pi_\theta,\,A_{i,t}\right)$ on unmasked rollouts, with $\pi_{\text{old}} = \pi_{\hat{\theta}_n}$ \hfill (cf.\ Eq.~\ref{eq: grpo_obj})

            \State $\mathbf{m}, \vsigma \gets \operatorname{IVON.step}(\nabla_\theta \mathcal{L})$ \Comment{Update posterior}
            \EndFor
        \end{algorithmic}
    \end{tcolorbox}
    \vspace{-0.5em}
    \captionof{alg}{\textbf{Chunked noising approach (\interleaved{}).} Rollouts are gathered in a buffer $\mathcal{R}$ across $N$ independent weight draws $\hat{\theta}_n$, each generating $G/N$ rollouts. The GRPO advantage is calculated on the accumulated buffer.}
    \label{alg: interleaved_prompt_level_noise}
\end{center}


\end{document}